\documentclass[style=google]{fairmeta}
\usepackage{mathptmx}% Times-like font
\usepackage{helvet}     % Helvetica
\usepackage{courier} 
\usepackage{xcolor}
\usepackage{enumitem}
\usepackage{float}
\usepackage{xltabular}
\usepackage{pgfplots}
\pgfplotsset{compat=1.18}
\usepackage{wrapfig}
\usepackage{svg}
\usepackage{tcolorbox}

\usepackage{devanagari}

\usepackage{array}
\usepackage{graphicx}
\usepackage{algorithm}
\usepackage{algpseudocode}
\usepackage{MnSymbol,wasysym}
\usepackage{tikzsymbols}

\usepackage{makecell}

\newcolumntype{P}[1]{>{\centering\arraybackslash}p{#1}}
\definecolor{firebrick}{RGB}{178,34,34}
\usepackage{verbatim}

\def\vtheta{{\bm{\theta}}}

\DeclareMathAlphabet{\mathsfit}{\encodingdefault}{\sfdefault}{m}{sl}
\SetMathAlphabet{\mathsfit}{bold}{\encodingdefault}{\sfdefault}{bx}{n}

\definecolor{lightgray}{gray}{0.95}
\lstdefinestyle{chatml}{
    backgroundcolor=\color{lightgray},
    basicstyle=\ttfamily,
    frame=none,
    columns=fullflexible,
    breaklines=true,
    keepspaces=true,
    showstringspaces=false
}

\title{\textsc{Param-1} BharatGen 2.9B Model
}

\author{Kundeshwar Pundalik}
\author{Piyush Sawarkar}
\author{Nihar Sahoo}
\author{Abhishek Shinde}
\author{Prateek Chanda}
\author{Vedant Goswami}

\author{Ajay Nagpal}
\author{Atul Singh}
\author{Viraj Thakur}
\author{Vijay Dewane}
\author{Aamod Thakur}
\author{Bhargav Patel}
\author{Smita Gautam}
\author{Bhagwan Panditi}
\author{Shyam Pawar}
\author{Madhav Kotcha}
\author{Suraj Racha}
\author{Saral Sureka}
\author{Pankaj Singh}
\author{Rishi Bal}
\author{Rohit Saluja}
\author{Ganesh Ramakrishnan}

\affiliation{BharatGen Team}

\abstract{
Large Language Models (LLMs) have emerged as powerful general-purpose reasoning systems, yet their development remains dominated by English-centric data, architectures, and optimization paradigms. This exclusionary design results in structural under-representation of linguistically diverse regions such as India, where over 20 official languages and 100+ dialects coexist alongside phenomena like code-switching and diglossia. We introduce \textsc{Param-1}, a 2.9B parameter decoder-only, text-only language model trained from scratch with an explicit architectural and linguistic focus on Indian diversity. \textsc{Param-1} is trained on a bilingual dataset consisting of only Hindi and English, constructed with a strong focus on fact-rich, high-quality content. It is guided by three core principles: equitable representation of Indic languages through a 25\% corpus allocation; tokenization fairness via a SentencePiece tokenizer adapted to Indian morphological structures; and culturally aligned evaluation benchmarks across IndicQA, code-mixed reasoning, and socio-linguistic robustness tasks. By embedding diversity at the pretraining level—rather than deferring it to post-hoc alignment—\textsc{Param-1} offers a design-first blueprint for equitable foundation modeling. Our results demonstrate that it serves as both a competent general-purpose model and a robust baseline for India-centric applications.}

\date{\today}
\correspondence{kundeshwar.pundalik@tihiitb.org}

\begin{document}

\maketitle

\section{Introduction}
\label{section:intro}

Large Language Models (LLMs) have emerged as the dominant computational paradigm for general-purpose reasoning, yet their current deployment reflects a narrow slice of the world's linguistic and cultural realities. Models such as GPT-4 and LLaMA  are typically trained on massive English-centric datasets, with tokenization and optimization strategies implicitly favoring high-resource, structurally similar languages. This leads to a structural asymmetry: while such models are universal in capability, they are not universal in inclusivity. Nowhere is this mismatch more pronounced than in the context of India—a linguistically plural, socio-culturally dense region encompassing over 20 official languages, 100+ regional dialects, and widespread phenomena such as code-switching and diglossia.

This paper introduces \textsc{Param-1}, a 2.9B parameter foundation model trained from scratch with an architectural, linguistic, and representational focus on India. \textsc{Param-1} is motivated by three core desiderata:
\begin{enumerate}
    \item representation: to ensure linguistic equity by explicitly allocating 25\% of the training corpus to Indic languages across diverse scripts and domains;
    \item tokenization fairness: to avoid the vocabulary fragmentation of Indian words under Western-trained tokenizers, we design a multilingual SentencePiece-based tokenizer that captures both prefix-root and agglutinative patterns common in Indian morphologies; and
    \item  evaluation alignment: to benchmark downstream utility in India-relevant tasks, we curate a suite of IndicQA, code-mixed reasoning, and socio-linguistic robustness evaluations.
\end{enumerate}

In doing so, \textsc{Param-1} challenges the prevailing notion that regional representation can be deferred to post-training alignment or fine-tuning. Instead, it advances a design-first philosophy for LLMs that explicitly internalizes linguistic and demographic diversity in the model’s very foundation. Through a combination of inclusive corpus design, culturally aware tokenization, and rigorous evaluation across Indic benchmarks, \textsc{Param-1} serves both as a performant general-purpose LLM and a proof-of-concept for equitable foundation modeling in underrepresented regions.
\subsection{Rethinking Language Models for India}
India’s rich linguistic and cultural diversity poses unique challenges and opportunities in the development of large language models (LLMs). With over 1.4 billion people and hundreds of languages and dialects spoken across its vast geography, India remains grossly underrepresented in mainstream AI systems. Most existing LLMs are trained predominantly on English or high-resource Western languages, rendering them ineffective when applied to Indian languages, contexts, or societal nuances. For instance, models like Meta’s LLaMA \cite{grattafiori2024llama} allocate a mere 0.01\% of training data to Indic languages—an alarming disparity for a region that constitutes nearly 18\% of the global population.

This imbalance manifests in poor comprehension, cultural misalignment, and biased outputs when such models are applied in Indian settings. The structural complexity of Indic languages, their rich oral traditions, and widespread code-mixing further exacerbate this gap. Building a foundational model that understands and reflects the lived realities of Indian users requires more than superficial fine-tuning—it demands architectural, linguistic, and dataset-level rethinking from the ground up.

\subsection{\textsc{Param-1}: A Ground-Up Model for India}

We present \textsc{Param-1}, a 2.9B parameter decoder-only language model trained from scratch with a design philosophy centered on Indian linguistic diversity. \textsc{Param-1} departs from conventional English-first scaling approaches and instead foregrounds equitable representation across major Indian language families (e.g., Indo-Aryan, Dravidian). Notably, the training corpus for \textsc{Param-1} includes over 25\% Indic language content—a stark contrast to prevailing models that marginalize Indic data below perceptual thresholds.

This intentional data construction is coupled with a tokenizer explicitly adapted to high-entropy, morphologically rich Indian scripts, enabling more faithful subword coverage across languages such as Hindi, Tamil, Telugu, Marathi, Bengali, and others. Training proceeds from scratch on a curated multilingual corpus that spans literary, governmental, scientific, and community-generated text, ensuring that the learned representations capture both formal and colloquial registers.

\textsc{Param-1} sets a new precedent in India-centric language modeling: not as a monolithic benchmark leader, but as a modular, transparent, and scalable foundation for downstream alignment. In the sections that follow, we describe the dataset composition, tokenizer design, training methodology, and evaluation results that collectively make \textsc{Param-1} a new baseline for India-centric language modeling.
\begin{figure}[H]
     \centering
    \includegraphics[width=0.8\linewidth]{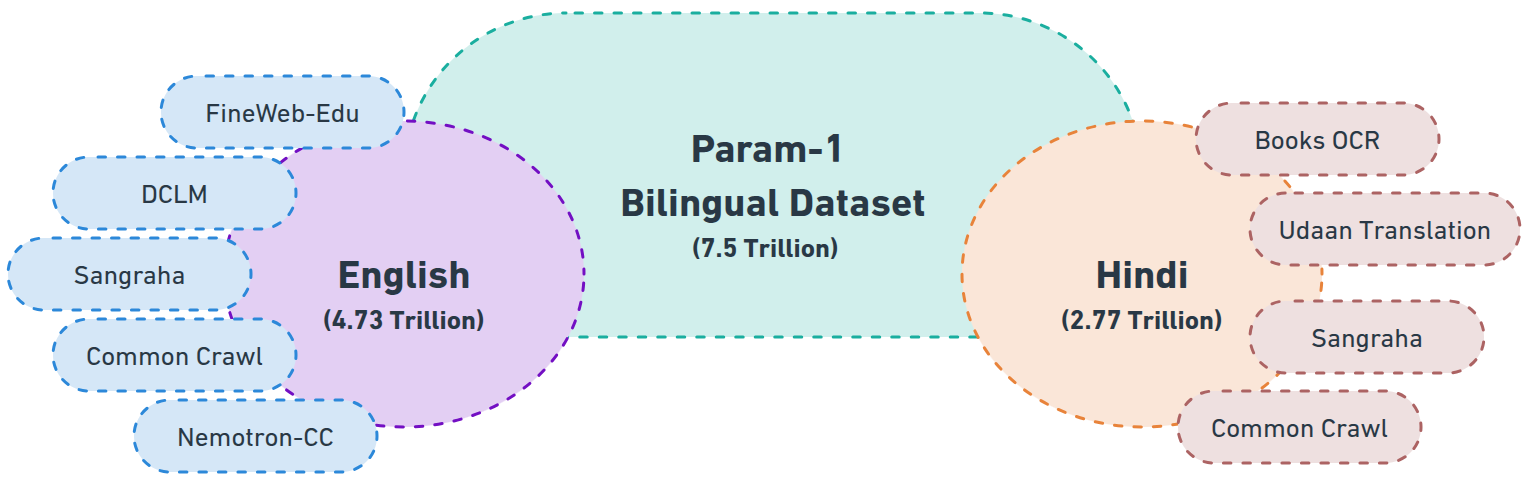}
    \caption{Pretraining Data Mixture}
    \label{fig:data}
\end{figure}

\section{Data Collection }

\subsection{Data sources and mixtures}
To train \textsc{Param-1}, we curated a massive 5 trillion-token multilingual dataset with a deliberate focus on India’s linguistic landscape—an approach that starkly contrasts with existing models where Indic data is almost absent. While 3.48 trillion tokens come from high-quality English corpora such as FineWeb-Edu, DCLM, Nemotron-CC, and filtered Common Crawl, the remaining 1.52 trillion tokens are composed of rich Hindi data sourced from Books OCR archives, government-funded Udaan translations, and web-scale resources cleaned such as Ai4bharat Sangraha dataset \cite{khan2024indicllmsuite}. This 25\% representation of Hindi alone makes \textsc{Param-1} uniquely positioned to model Indian linguistic nuances at scale. The dataset covers diverse content types—literary, instructional, conversational, and informal—and underwent meticulous preprocessing, including deduplication, noise reduction, and Indic-aware normalization. Figure\ref{fig:data} illustrates our dual-source strategy that enables \textsc{Param-1} to operate fluently across English and Indic contexts, setting a new benchmark for culturally grounded, multilingual LLMs.

\subsection{Data Curation and filtering}
To ensure high-quality and reliable training data, we employed a multi-stage data curation and filtering pipeline using the NVIDIA NeMo Curator framework. This pipeline combined classifier-based, heuristic, and rule-based filtering techniques to systematically clean and structure both English and Indic datasets before model pretraining.
\subsubsection{Classifier and Heuristic Filtering}
We applied two primary types of document-level filtering:
\begin{itemize}
    \item \textbf{Classifier-based Filtering:} We used pre-trained and custom-trained fastText models to score each document on overall quality. Based on these scores, documents were classified into low, medium, and high quality buckets. Only high-quality documents were retained for downstream training.
    \item \textbf{Heuristic Filtering:} Complementing the classifiers, we used rule-based filters from NeMo such as WordCountFilter and MeanWordLengthFilter to remove noisy, malformed, or overly short/long documents. These filters targeted documents with poor linguistic structure, abnormal word length distributions, or insufficient content.
\end{itemize}

\subsection{Language Identification and Unicode Normalization}
We ensured accurate multilingual processing through two additional preprocessing steps:
\begin{itemize}
    \item \textbf{Language Identification:} Using the \textit{FastTextLangId}\cite{cui2020language} filter, we detected the language of each document and separated corpora by language with high confidence. This was required to avoid the problem of code mixing.
    \item \textbf{Unicode Reformatting:} Documents containing improperly encoded characters were processed using the \textit{UnicodeReformatter} class, which internally leverages the \textit{ftfy} library to fix broken Unicode sequences and improve textual consistency.
\end{itemize}

\subsection{Deduplication}
To eliminate redundant samples and reduce training inefficiency, we applied both:

\begin{itemize}
    \item \textbf{Exact Deduplication:} Identical documents (bitwise duplicates) were collapsed to a single copy.
    \item \textbf{Fuzzy Deduplication:} We employed GPU-accelerated algorithms to remove near-duplicate texts using similarity heuristics. This step was crucial in large web-scale corpora where minor template variations often occur.
\end{itemize}
\subsection{PII Detection and Removal}
We applied the PiiModifier utility to identify and redact Personally Identifiable Information (PII) such as names, addresses, emails, and phone numbers. This step was essential to uphold user privacy and ensure compliance with data protection norms.

\subsection{Code and Math removal}
To ensure that the pretraining corpus for \textsc{Param-1} remains focused on natural language understanding and generation, we implemented a filtration strategy to systematically remove documents containing code snippets and mathematical expressions.

Our motivation was twofold: (i) \textsc{Param-1} is designed as a text-only model optimized for bilingual Hindi-English usage in general-purpose and culturally grounded tasks rather than code generation, and (ii) inclusion of code/math-heavy documents can skew token distribution and reduce linguistic diversity.

The filtration pipeline involved the following steps:

\begin{itemize}
\item \textbf{Regex-based filtering:} We applied regular expressions to detect and exclude documents containing common code patterns (e.g., function definitions, import statements, angle-bracket tags, or syntax resembling Python, Java, HTML, etc.).

\item \textbf{Math expression detection:} We removed documents with a high frequency of LaTeX-like or Unicode math symbols (e.g., \texttt{frac}, \texttt{sum}, $\int$, $\sqrt{\cdot}$, $\sum$)
 or heavy numerical token density, which often indicates formula-heavy content.

\item \textbf{Heuristic scoring:} Documents were scored for code/math likelihood using shallow classifiers trained on a small manually labeled subset. Documents scoring above a defined threshold were discarded.

\item \textbf{Structural cues:} We filtered out files containing structured formatting cues (e.g., consistent indentation patterns, inline code blocks, or markdown cells) commonly found in scraped code or educational repositories.
\end{itemize}

\section{Tokenizer}
\label{section:tokenizer}
For \textsc{Param-1}, we employ a customized tokenizer trained using the SentencePiece BPE algorithm on an in-house curated corpus spanning diverse Indian languages and domains. We adopt a vocabulary size of 128K tokens, chosen after extensive ablation studies to strike a balance between tokenization efficiency and script coverage. To improve handling of rare characters, particularly in low-resource languages, we explicitly included all unique script symbols before training, avoiding over-fragmentation via byte fallback. The tokenizer also integrates a byte fallback mechanism, ensuring robustness across unknown symbols and non-Indic text. Furthermore, a pre-tokenization layer splits digits and whitespace patterns, aiding model performance in arithmetic and programming tasks. This tokenizer design ensures compact and semantically coherent token sequences across India’s multilingual landscape.

\begin{table*}[h]
    \centering
    \small
    \resizebox{\textwidth}{!}{%
    \begin{tabular}{lccccccc}
    \toprule
    \bf Language &
    \bf BharatGen-64K v1 &
    \bf BharatGen-128K v1 &
    \bf Qwen &
    \bf LLaMA &
    \bf Nemotron\_Mistral &
    \bf Nemotron\_Mini &
    \bf Sarvam-M \\
    \midrule
    asm – Assamese & 2.10 & 1.83 & 7.18 & 8.06 & 4.24 & 4.58 & 4.24 \\
    ben – Bengali & 2.04 & 1.78 & 6.92 & 7.85 & 2.93 & 2.65 & 2.93 \\
    eng – English & 1.65 & 1.51 & 1.36 & 1.35 & 1.37 & 1.35 & 1.37 \\
    guj – Gujarati & 2.08 & 1.83 & 8.53 & 9.54 & 3.59 & 15.17 & 3.59 \\
    hin – Hindi & 1.58 & 1.43 & 4.66 & 2.65 & 1.97 & 1.77 & 1.97 \\
    kan – Kannada & 2.59 & 2.17 & 11.08 & 13.81 & 3.82 & 4.02 & 3.82 \\
    mai – Maithili & 1.93 & 1.74 & 4.67 & 2.85 & 2.53 & 2.28 & 2.53 \\
    mal – Malayalam & 3.18 & 2.67 & 13.30 & 16.00 & 4.88 & 4.71 & 4.88 \\
    mar – Marathi & 2.03 & 1.76 & 6.46 & 3.86 & 3.14 & 2.62 & 3.14 \\
    nep – Nepali & 1.87 & 1.60 & 6.28 & 3.61 & 3.04 & 2.32 & 3.04 \\
    ori – Odia & 2.06 & 1.75 & 12.92 & 15.91 & 17.23 & 17.24 & 17.23 \\
    pan – Punjabi & 1.88 & 1.64 & 7.39 & 7.88 & 3.12 & 12.70 & 3.12 \\
    san – Sanskrit & 3.28 & 2.99 & 8.00 & 4.75 & 4.26 & 4.32 & 4.26 \\
    snd – Sindhi & 1.74 & 1.54 & 3.09 & 2.99 & 2.65 & 2.83 & 2.65 \\
    tam – Tamil & 2.54 & 2.18 & 9.75 & 11.89 & 3.71 & 3.57 & 3.71 \\
    tel – Telugu & 2.81 & 2.35 & 11.45 & 13.30 & 3.90 & 3.77 & 3.90 \\
    \bottomrule
    \end{tabular}
    }
    \caption{Tokenizer Fertility score with Indian languages. Lower is better.}
    \label{table:language_eval}
\end{table*}

To evaluate the effectiveness of our tokenizer, we compared its fertility scores—defined as the average number of tokens generated per word—against a range of other prominent tokenizers including those from Qwen \cite{bai2023qwen}, LLaMA\cite{grattafiori2024llama}, Nemotron-Mistral \cite{karamcheti2021mistral}, Nemotron-Mini \cite{joshi2024adapting}, Sarvam-M, and two configurations of BharatGen (64K and 128K). As illustrated in Table \ref{table:language_eval}, our BharatGen-128K v1 tokenizer consistently achieves lower fertility scores across multiple Indian languages such as Hindi, Bengali, Tamil, and Gujarati, indicating more compact and semantically aligned tokenization. In contrast, tokenizers like LLaMA and Qwen exhibit significantly higher fertility—particularly for Indic scripts—highlighting their inefficiencies in tokenizing Indian language content. The performance gap is most prominent in languages like Odia, Kannada, and Malayalam, where our tokenizer's script-aware design and character-level vocabulary coverage result in reduced fragmentation and improved efficiency. These results underscore the importance of domain-specific tokenizer design when targeting linguistically diverse environments like India. The tokenizer mentioned in above refers to the Bharatgen in-house tokenizer; however, \textsc{Param-1} was trained using the Nemotron tokenizer \cite{parmar2024nemotron}.

\subsection{Domain-Aware Corpus Construction}

To ensure strong performance beyond natural language, our tokenizer training data includes curated domain corpora:
\begin{itemize}
    \item \textbf{Programming:} We sample from the StarCoder dataset, which includes code from over 80 programming languages. This enables robust handling of identifiers, syntax tokens, and indentation structures.
    \item \textbf{Mathematics:} We integrate OpenWebMath and LaTeX-Formulas datasets, combining structured equations, LaTeX markup, and word problems (e.g., GSM8K, SVAMP). Byte fallback ensures graceful degradation for rarely used symbols.
\end{itemize}

\subsection{Fertility-Aware Language Mixture Optimization}

To fairly represent each Indian language during tokenizer training, we propose a fertility-driven data sampling strategy based on momentum-weighted allocation. Let $f_l^N$ denote the fertility score (average tokens per word) for language $l$ at iteration $N$. Lower fertility implies more compact, efficient tokenization. Our iterative rebalancing algorithm adjusts the character-level mixture $m_l^N$ across training steps to minimize fertility divergence from the ideal score (set to 1.0). The core update rule is:
\begin{align*}
\delta_l^N &= \frac{f_l^N - f_{\text{best}}}{f_{\text{range}}^N}, \\
w_l^N &= \delta_l^N + \epsilon, \quad
t_l^N = \frac{w_l^N}{\sum_k w_k^N}, \\
m_l^N &= (1 - \mu) \cdot m_l^{N-1} + \mu \cdot t_l^N, \\
C_l^N &= \text{round}(m_l^N \cdot T),
\end{align*}

\noindent where $\mu$ is a momentum factor, $\epsilon$ is a smoothing constant, and $T$ is the total number of characters per iteration. This adaptive strategy ensures that over-fragmented languages (e.g., Malayalam) are emphasized during training, leading to improved script-level coverage and lower token redundancy.

\section{Model and Architecture}
\label{section:methods}
The transformer architecture has become the foundation for most state-of-the-art language models, offering an effective structure for learning patterns in large-scale textual data. Among its various forms, the decoder-only configuration has proven especially efficient for tasks involving text generation, such as dialogue systems, summarization, and code generation. 

\textsc{Param-1} follows a standard decoder-only dense transformer architecture, similar to widely adopted designs like GPT and LLaMA. It consists exclusively of transformer blocks with masked self-attention, optimized for autoregressive language modeling. The model uses a single stack of transformer layers without any mixture-of-experts or encoder components, focusing on simplicity, stability, and inference efficiency. Key architectural components include multi-head self-attention with rotary positional embeddings, layer normalization, and feedforward MLP layers with SwiGLU \cite{shazeer2020glu} activation. \textsc{Param-1} is trained with a maximum sequence length of 2048 tokens, and uses a custom tokenizer optimized for both English and Indic languages to reduce token-to-word inflation, especially for underrepresented scripts.

\begin{table*}[h]
    \centering
    \small
    \begin{tabular}{l @{\hskip 2em} >{\centering\arraybackslash}p{5cm}}
    \toprule
    \bf Architecture attributes & \bf Values \\
    \midrule
    \texttt{Model Architecture}               & \texttt{causal-language-model} \\
    \texttt{Hidden size}              & \texttt{2048} \\
    \texttt{Intermediate size}        & \texttt{7168} \\
    \texttt{Max Position Embeddings} & \texttt{2048} \\
    \texttt{Num of Attention Heads} & \texttt{16} \\
    \texttt{Rope theta}               & \texttt{10000} \\
    \texttt{Num of Hidden Layers}   & \texttt{32} \\
    \texttt{Num of Key Value Heads} & \texttt{8} \\
    \texttt{Activation Function}       & \texttt{fast-swiglu} \\
    \texttt{Attention Type}                 & \texttt{Grouped-query attention} \\
    \texttt{Precision}                 & \texttt{bf16-mixed} \\
    \bottomrule
    \end{tabular}
    \caption{Architecture Details of \textsc{Param-1}}
    \label{table:architecture}
\end{table*}

\section{Training}
Here we detail out our training strategy utilised for training our \textsc{Param-1} model.

We begin by formally defining the pretraining procedure. Under an autoregressive next-token prediction framework, let the input be a token sequence \( (x_1, x_2, \dotsc, x_T) \). The model, parameterized by \( \vtheta \), specifies a conditional probability distribution \( p_\vtheta(x_t \mid x_{<t}) \) at each time step \( t \), where \( x_{<t} = (x_1, \dotsc, x_{t-1}) \) represents the sequence of preceding tokens. The \emph{next-token cross-entropy loss} over the sequence is defined as:
\[
\mathcal{L}_{\mathrm{CE}}(\vtheta) = -\sum_{t=1}^{T} \log p_\vtheta(x_t \mid x_{<t})
\]
The goal is to minimize over a corpus of tokens the above cross entropy loss 

\subsection{Pre-Training}
To ensure stable convergence and effective generalization, we trained \textsc{Param-1} in three progressive phases. Each phase was designed with a distinct objective: data bootstrapping, factual preservation, and long-context adaptation. This structured approach allowed the model to gradually acquire linguistic and structural competence across diverse domains and sequence lengths. 

\subsubsection{PT Phase-1: Bootstrap Training}
The first phase focused on initializing core language understanding using high-quality, diverse text. Training was conducted on a 5 trillion-token multilingual corpus, comprising 3.48 trillion English tokens and 1.52 trillion Hindi tokens, carefully curated from sources such as FineWeb-Edu \cite{penedo2024fineweb}, DCLM \cite{tissenbaum2017dclm}, Nemotron-CC\cite{su2024nemotron}, Sangraha, Books OCR, and Udaan \cite{maheshwari2023udaan} \ref{fig:data}. This phase was executed on a 64-node SLURM-managed cluster, with each node equipped with 8× NVIDIA H100 GPUs, leveraging full data, tensor, and pipeline parallelism via the NeMo framework and Megatron backend \cite{shoeybi2019megatron}.

\begin{figure}[H]
     \centering
    \includegraphics[width=0.5\linewidth]{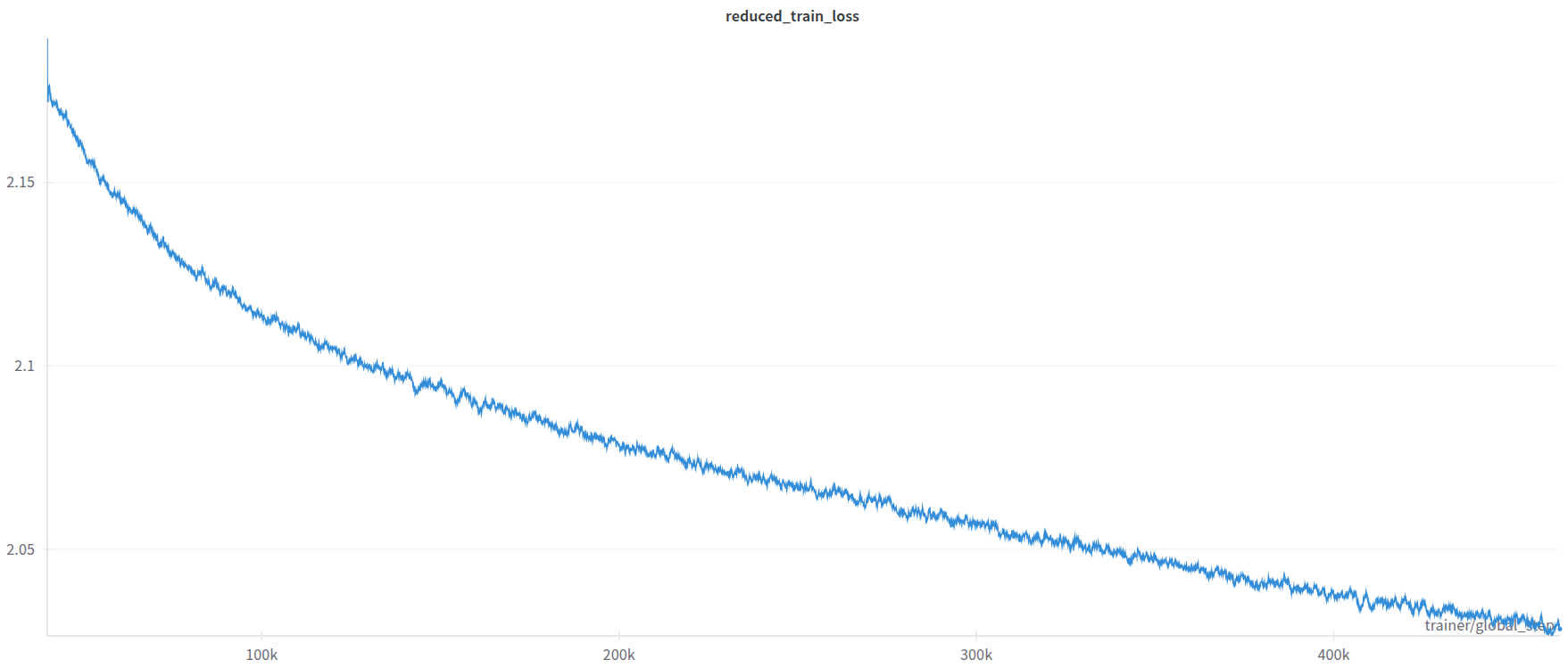}
    \caption{Pre-training loss curve}
    \label{fig:pt-loss}
\end{figure}

Training was restricted to short-to-medium length sequences (up to 2,048 tokens), allowing the model to stably learn foundational syntactic, semantic, and cross-lingual structures. Conservative optimization settings—such as a low initial learning rate and smaller batch sizes—were adopted to ensure stable convergence. This phase laid the groundwork for building the model’s general language competence and multilingual robustness before moving to more specialized objectives. Figure \ref{fig:pt-loss} depicts training loss during pre-training.

\subsubsection{PT Phase-2: Factual Preservation}
During inference evaluations after Phase 1, we observed that the model occasionally struggled with recalling factual information, particularly in response to knowledge-intensive prompts. To address this, Phase 2 was dedicated to enhancing the model’s factual consistency and knowledge retention through targeted curriculum training.

This phase was trained on a 2 trillion-token corpus, evenly split between Hindi and English (1T each). The dataset was built with a strong emphasis on fact-rich content. Specifically, 20\% of the data was drawn from the original Phase 1 corpus (PT1) to maintain linguistic stability, while the remaining 80\% comprised newly curated data with following split \ref{fig:phase-2}:
\begin{wrapfigure}{l}{0.4\linewidth}
  \centering
  \includegraphics[width=\linewidth]{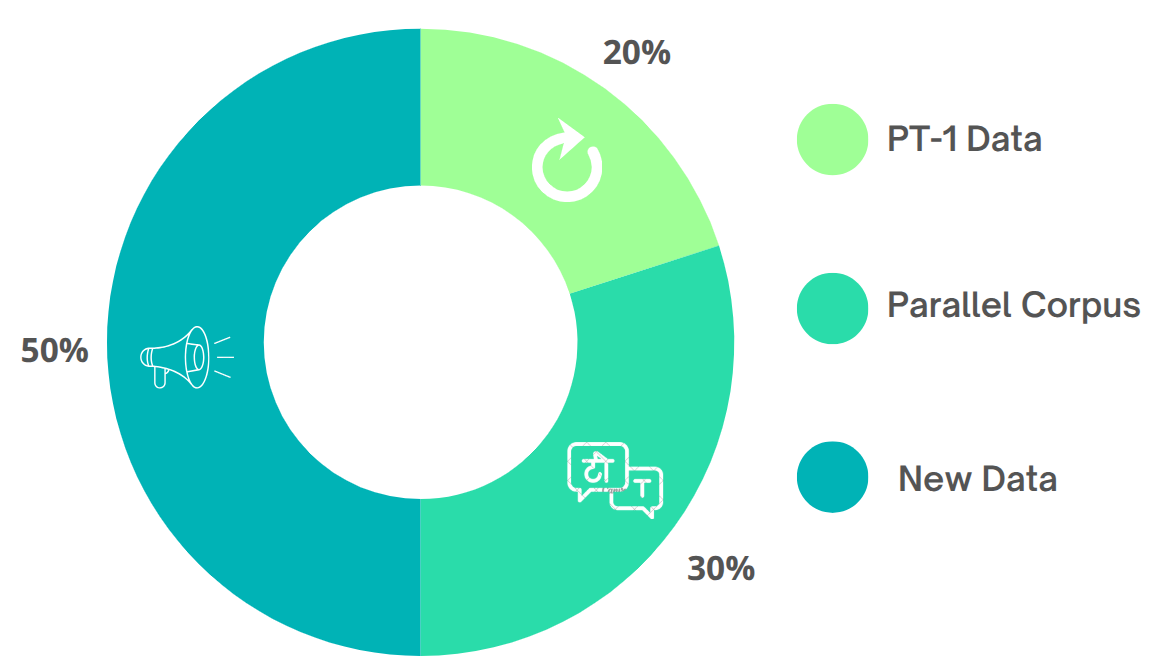}
  \caption{Pre-Training Phase-2 data mixture}
  \label{fig:phase-2}
\end{wrapfigure}
\begin{itemize}
    \item 30\% Parallel Corpus: Sentence-aligned English-Hindi pairs from high-quality translation datasets (e.g., Udaan \cite{maheshwari2024dictdis} , Samanantar \cite{ramesh2022samanantar}), allowing the model to strengthen cross-lingual factual grounding.
    \item 50\% Monolingual New Data: Fresh English and Hindi documents from domains such as encyclopedias, textbooks, technical manuals, and verified online resources.
\end{itemize}

Training was performed on 32 nodes, each equipped with 8× NVIDIA H100 GPUs, utilizing full model parallelism and mixed-precision optimization. The model continued to use a 2048-token sequence length, but the sampling strategy was adjusted to upweight high-information content and discourage repetition or hallucination.

This phase significantly improved the model’s factual fluency and its ability to produce consistent and verifiable outputs across both English and Indic queries.

\subsubsection{PT Phase-3: Long-Context Adaptation}
In the final phase of training, we focused on enabling long-context understanding and retention, which is essential for tasks involving multi-hop reasoning, document-level comprehension, and long-form generation.

\definecolor{barA}{RGB}{31,119,180}
\definecolor{barB}{RGB}{255,127,14}
\definecolor{barC}{RGB}{44,160,44}
\definecolor{barD}{RGB}{214,39,40}
\definecolor{barE}{RGB}{148,103,189}
\definecolor{barF}{RGB}{140,86,75}
\definecolor{barG}{RGB}{227,119,194}
\definecolor{barH}{RGB}{127,127,127}

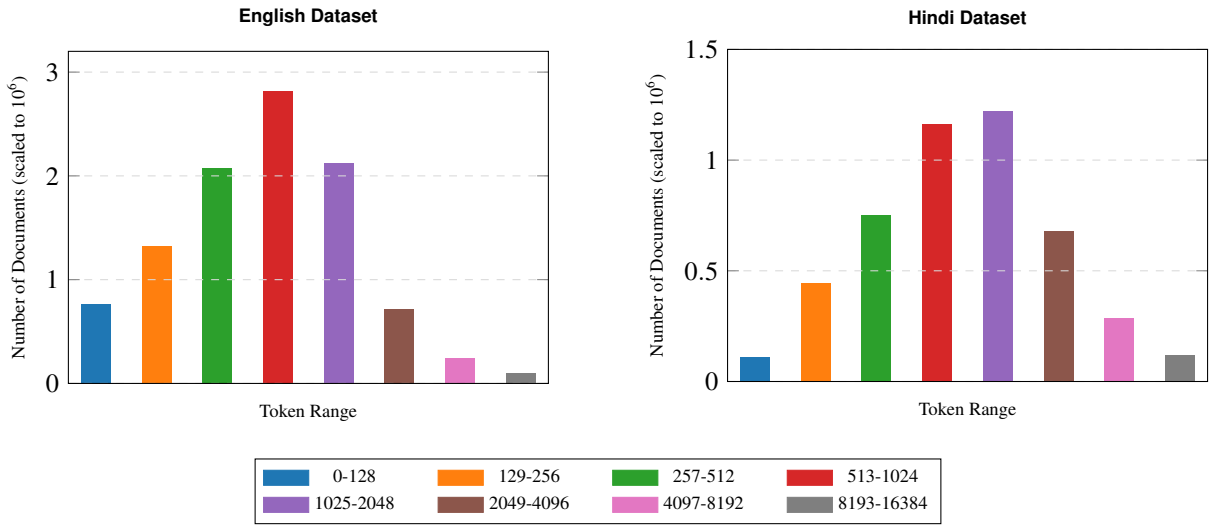
\begin{figure}[htbp]
\centering

\begin{minipage}{0.48\textwidth}
\centering
\begin{tikzpicture}
\begin{axis}[
    width=\textwidth,
    height=6cm,
    ybar,
    bar width=11pt,
    ymin=0,
    ymax=3.2,
    ytick={0,1,2,3},
    ylabel={\scriptsize Number of Documents (scaled to $10^6$)},
    xlabel={\scriptsize Token Range},
    xtick=\empty,
    xticklabel style={rotate=45, anchor=east, font=\scriptsize},
    symbolic x coords={0-128,129-256,257-512,513-1024,1025-2048,2049-4096,4097-8192,8193-16384},
    enlarge x limits=0.80,
    axis on top,
    grid=both,
    grid style={dashed, gray!30},
    title={\scriptsize \textbf{English Dataset}},
    legend style={draw=none}, % suppress legend here
    ]
    \addplot+[ybar, fill=barA, draw=none] coordinates {(0-128,0.768692)}; 
    \addplot+[ybar, fill=barB, draw=none] coordinates {(129-256,1.319696)};
    \addplot+[ybar, fill=barC, draw=none] coordinates {(257-512,2.079729)};
    \addplot+[ybar, fill=barD, draw=none] coordinates {(513-1024,2.815809)};
    \addplot+[ybar, fill=barE, draw=none] coordinates {(1025-2048,2.124629)};
    \addplot+[ybar, fill=barF, draw=none] coordinates {(2049-4096,0.717496)};
    \addplot+[ybar, fill=barG, draw=none] coordinates {(4097-8192,0.247746)};
    \addplot+[ybar, fill=barH, draw=none] coordinates {(8193-16384,0.098551)};
\end{axis}
\end{tikzpicture}
\end{minipage}
\hfill
\begin{minipage}{0.48\textwidth}
\centering
\begin{tikzpicture}
\begin{axis}[
    width=\textwidth,
    height=6cm,
    ybar,
    bar width=11pt,
    ymin=0,
    ymax=1.5,
    ytick={0,0.5,1,1.5},
    ylabel={\scriptsize Number of Documents (scaled to $10^6$)},
    xlabel={\scriptsize Token Range},
    xtick=\empty,
    xticklabel style={rotate=45, anchor=east, font=\scriptsize},
    symbolic x coords={0-128,129-256,257-512,513-1024,1025-2048,2049-4096,4097-8192,8193-16384},
    enlarge x limits=0.80,
    axis on top,
    grid=both,
    grid style={dashed, gray!30},
    title={\scriptsize \textbf{Hindi Dataset}},
    legend style={draw=none}, % suppress legend here
    ]
    \addplot+[ybar, fill=barA, draw=none] coordinates {(0-128,0.110697)};
    \addplot+[ybar, fill=barB, draw=none] coordinates {(129-256,0.443936)};
    \addplot+[ybar, fill=barC, draw=none] coordinates {(257-512,0.754026)};
    \addplot+[ybar, fill=barD, draw=none] coordinates {(513-1024,1.161479)};
    \addplot+[ybar, fill=barE, draw=none] coordinates {(1025-2048,1.220583)};
    \addplot+[ybar, fill=barF, draw=none] coordinates {(2049-4096,0.680297)};
    \addplot+[ybar, fill=barG, draw=none] coordinates {(4097-8192,0.287794)};
    \addplot+[ybar, fill=barH, draw=none] coordinates {(8193-16384,0.117519)};
\end{axis}
\end{tikzpicture}
\end{minipage}

% Separate legend below the two plots
\vspace{1em}
\begin{tikzpicture}
\begin{axis}[
    hide axis,
    xmin=0, xmax=1,
    ymin=0, ymax=1,
    legend columns=4,
    legend style={
        at={(0.5,0)},
        anchor=north,
        font=\scriptsize,
        /tikz/every even column/.append style={column sep=0.5cm}
    },
]
% Add legend entries only (no plots)
\addlegendimage{area legend, ybar, fill=barA, draw=barA}
\addlegendentry{0-128}
\addlegendimage{area legend, ybar, fill=barB, draw=barB}
\addlegendentry{129-256}
\addlegendimage{area legend, ybar, fill=barC, draw=barC}
\addlegendentry{257-512}
\addlegendimage{area legend, ybar, fill=barD, draw=barD}
\addlegendentry{513-1024}
\addlegendimage{area legend, ybar, fill=barE, draw=barE}
\addlegendentry{1025-2048}
\addlegendimage{area legend, ybar, fill=barF, draw=barF}
\addlegendentry{2049-4096}
\addlegendimage{area legend, ybar, fill=barG, draw=barG}
\addlegendentry{4097-8192}
\addlegendimage{area legend, ybar, fill=barH, draw=barH}
\addlegendentry{8193-16384}
\end{axis}
\end{tikzpicture}

\caption{Document Length Distribution by token range. Each color corresponds to a token range (see legend).}
\label{fig:token-dist}
\end{figure}

As shown in Figure \ref{fig:token-dist}, the document length distribution in our training corpus was diverse, with a significant number of documents falling in the 512–1024 and 1025–2048 token ranges—ensuring a strong baseline of medium-length examples. To support long-context adaptation, we incorporated a substantial number of documents with lengths exceeding 2,048 tokens, including over 70K documents in the 2K–4K range, ~250K between 4K–8K, and nearly 100K exceeding 8K tokens. This distribution was essential in helping the model learn to handle long-range dependencies and maintain coherence across extended inputs, particularly for document-level tasks such as summarization, QA, and RAG-style retrieval settings.

We trained this phase on a 500 billion-token dataset, equally split between English (250B) and Hindi (250B). The dataset was constructed with the following composition:
\begin{itemize}
    \item 20\% reused data from earlier pretraining phases (PT1 + PT2), ensuring continuity in foundational language patterns.
    \item 30\% high-quality parallel corpus (sentence-aligned English-Hindi pairs), enabling the model to maintain alignment across longer bilingual spans.
    \item 50\% newly sourced monolingual data, consisting of long-form documents such as books, articles, technical manuals, and multi-paragraph conversational data to simulate realistic long-context scenarios.
\end{itemize}

Training was conducted on 32 nodes, each with 8× NVIDIA H100 GPUs, leveraging efficient pipeline and tensor parallelism configurations. We also adjusted the sampling and sequence packing strategies to favor dense, information-rich content while minimizing empty context segments. This phase significantly enhanced the model’s ability to maintain coherence, reference earlier parts of a prompt, and generate context-aware responses across longer sequences.

\subsection{Post Training}
To align \textsc{\textsc{Param-1}} with a broad range of downstream tasks and improve its effectiveness in real-world, interactive, and application-driven settings, we conducted a dedicated phase of instruction fine-tuning. This phase enables the bilingual model, trained exclusively on English and Hindi, to follow natural language prompts and generate coherent and contextually grounded outputs. While the primary focus is on Indian linguistic and domain contexts, care has been taken to ensure the model remains applicable to general-purpose, globally relevant use cases as well.

By fine-tuning on a carefully curated set of instruction–response pairs spanning diverse domains, communicative styles, and formality levels in both English and Hindi, \textsc{Param-1} internalize domain-specific tasks, sociolinguistic norms, and culturally-grounded usage patterns and aligns with real-world use cases typical in Indian governance, history, education, legal, agriculture, and public services.

\subsubsection{Dataset Curation and Filtering Strategy}
To ensure relevance, inclusiveness, and performance across Indian use cases, we adopted a \textbf{hybrid dataset development} strategy by combining curated open-source data with task-specific synthetic and crowdsourced annotations.  However, to ensure the final dataset was clean, safe, and aligned with the goals of instruction tuning, we applied a rigorous three-phase strategy: \textbf{collection}, \textbf{filtering}, and \textbf{quality scoring}.
\begin{enumerate}[label=(\alph*), left=0em, labelsep=1em]
    \item \textbf{Data Sources} \\ 
Initially, we gathered a list of instruction fine-tuning datasets, either natively available in English or Hindi. We began by collecting publicly available instruction tuning datasets such as \textit{NATURALINSTRUCTIONS \cite{super-natural-instructions}, OpenOrca \cite{OpenOrca}, UltraChat \cite{ultrachat}, IndicAlign \cite{khan2024indicllmsuite}, Nemotron SFT \cite{bercovich2025llamanemotronefficientreasoningmodels}, Dolly \cite{DatabricksBlog2023DollyV2}, UltraFeedback \cite{cui2023ultrafeedback}}, among others. While most of these resources were originally in English, we were also able to incorporate Hindi datasets like \textit{IndicAlign}, \textit{Dharampal Book QA}, and \textit{IndicQA} \cite{singh2025indicqabenchmarkmultilingual}. Additionally, we curated a collection of digitized and OCR-processed bilingual books and materials from Indian civil services preparation resources. These were subsequently reformulated into instruction–response pairs in both English and Hindi to augment the bilingual fine-tuning corpus.
    \item \textbf{Two-Stage Filtering Pipeline}\\
To ensure that only useful and safe instruction data was included, we applied a two-stage filtering process:

\begin{enumerate}[label=\textbf{Stage \arabic*:}, leftmargin=0pt, itemindent=!, labelsep=1em, align=left]

\item \textbf{Domain and Safety Filtering}
\begin{itemize}
    \item \textbf{Code and Math Removal:} Examples involving programming, LaTeX, or maths, or any other symbolic computation were excluded, as our focus is on general-purpose question-answering and conversational fluency. We employ both model-based and rule-based filtering to remove such content from the curated fine-tuning corpus.

    \item \textbf{Safety and Toxicity Filtering:} Both rule-based heuristics and model-assisted classifiers were applied to detect and remove content that was offensive, casteist, or politically or religiously sensitive in the Indian context. Special attention was given to Hindi samples due to higher lexical ambiguity in low-resource safety detection.
\end{itemize}

\item \textbf{Instruction Quality Scoring} \\
Remaining instruction–response samples that passed the initial domain and safety filtering were subsequently evaluated using the \textit{Qwen-32B-IT} model, employed as a reference scorer. Each example was rated according to a comprehensive rubric designed to ensure instructional quality, linguistic accuracy, and cultural relevance. The rubric consisted of the following four criteria:

\begin{itemize}
\item \textbf{Relevance:} Assesses whether the instruction is appropriate and meaningful in the context of Indian-specific tasks (e.g., governance, education, public policy) or broadly applicable general knowledge domains. Instructions that lacked contextual grounding or appeared artificially constructed were down-rated.

\item \textbf{Fluency:} Measures the grammaticality, syntactic correctness, and idiomatic usage in both Hindi and English. This criterion was especially critical for ensuring the naturalness of Hindi prompts and responses, where literal translations often reduce interpretability.

\item \textbf{Helpfulness:} Evaluates the informativeness, precision, and utility of the response. Responses were required to be sufficiently detailed, free from hallucinations, and directly address the user's query or task intent.

\item \textbf{Prompt–Response Alignment:} Judges how accurately the response corresponds to the given instruction. The response was expected to remain faithful to the instruction’s scope, avoid digressions, and conclude the task logically and coherently.
\end{itemize}

Only those samples that achieved a perfect score of \textbf{5 out of 5} on all four dimensions were retained in the final dataset. This strict scoring protocol ensured a high-quality bilingual dataset, optimized for alignment, safety, and downstream usability.

Further details regarding the specific prompts, heuristic filters, and scoring templates used during both the filtering and scoring stages can be found in Appendix.

\end{enumerate}

    % We first removed examples that included code snippets or mathematical expressions, as the objective was to fine-tune for language understanding rather than programming or symbolic reasoning. Additionally, we applied toxicity and hate speech filtering using both keyword-based heuristics and model-assisted classifiers to eliminate unsafe or offensive content. The prompts and filters used during this stage are detailed in Appendix \ref{appendix:prompt}.

    \item \textbf{Tülu 3 Dataset Integration} \\
    To supplement our core corpus, we incorporated the \textbf{T\"ulu 3} dataset \cite{lambert2025tulu3pushingfrontiers}, known for its instruction diversity and coverage of reasoning and open-ended tasks. Although originally in English, we selected a subset of culturally neutral and broadly relevant instructions for inclusion.

As with the primary dataset, this subset was processed as follows:

\begin{itemize}
    \item All examples containing code, math, or unsafe content were excluded.
    \item All instances, not belonging to English or Hindi, were removed using rule-based language filtering.
    % \item A high-quality Hindi translation subset was generated using in-house neural translation systems, followed by manual review.
    \item Qwen-32B-IT scoring was applied to all bilingual samples, retaining only those with a score of 5.
\end{itemize}

This process yielded approximately \textbf{207,000 high-quality English–Hindi instruction–response pairs} from \textbf{T\"ulu 3}, which were merged into our final training pool.

    \item \textbf{BharatGen In-house Synthetic dataset} \\
    To capture India-specific linguistic and domain variation more comprehensively, we developed \textbf{BharatGen}, a bilingual instruction dataset tailored to Indian socio-cultural and institutional realities. Major sources included:

\begin{itemize}
    \item \textbf{DharmaWiki:} Curated entries explaining cultural, philosophical, and historical aspects of Sanatana Dharma, reformatted as instructional Q\&A in Hindi and English.

    \item \textbf{OCR-Processed Texts:} Indian literary texts and speeches in Hindi and bilingual formats were transformed into prompts for summarization, interpretation, and conversational tasks using \textit{DeepSeek-V3} \cite{liu2024deepseek}.

    \item \textbf{Spoken Language Transcripts:} Hindi-English code-switched text derived from text-to-speech applications, IVR simulations, and YouTube transcripts provided naturally conversational data in both languages.

    \item \textbf{Domain-Specific Instruction Sets:} Corpora from Indian legal, agricultural, financial, and governance domains were converted into instructional prompts and paired responses, reflecting real-world questions asked in English, Hindi, and Hinglish. 
    % in government portals, WhatsApp bots, and helpdesk
\end{itemize}

For all these, we create a synthetic instruction-response pair dataset of diverse task patterns such as \textit{open-ended QA, context-based QA, Yes/No QA, summarization, story generation, ACR-style QA, Squad-style QA, Question generation}, etc., by prompting \textit{DeepSeek-V3} \cite{deepseekai2025deepseekv3technicalreport}. More about these different task types is detailed in Appendix. This is followed by the quality-filtering step as described above to ensure that we retain only useful instruction-tuning pairs. This bilingual, grounded dataset formed the backbone of \textsc{Param-1}’s instruction tuning corpus.

    % To ensure only high-quality examples were retained, we used Mistral-7B \cite{jiang2023mistral7b} as a scoring model. Each instruction–response pair was evaluated and assigned a score from 1 to 5 based on task relevance, fluency, helpfulness, and prompt alignment. Only examples that received a score of 5 were retained. This process ensured the dataset was both clean and of instructional value.
\end{enumerate}

By aggregating all the above data sources, we curated two distinct supervised fine-tuning (SFT) datasets, referred to as \textsc{\textsc{Param-1}-473k} and \textsc{\textsc{Param-1}-1M}. The \textsc{\textsc{Param-1}-1M} dataset comprises approximately 1 million instruction–response pairs, representing the full breadth of our bilingual instruction corpus. In contrast, \textsc{\textsc{Param-1}-473k} is a high-quality, hard-filtered subset containing around 473,000 examples that meet the strictest thresholds for alignment, safety, and fluency. A comparative evaluation of these two datasets and their impact on model performance is presented in Section~\ref{sec:results}.

\subsubsection{Training Setup}
The instruction fine-tuning was conducted on top of the pre-trained \textsc{\textsc{Param-1}} checkpoint using a causal language modeling objective with masked prompt loss. The entire process was bilingual, with the training batches consisting of a 50:50 mix of English and Hindi instructions.

\paragraph{Training Configuration:}
\begin{itemize}
    \item \textbf{Sequence Format:} Each sample consisted of a natural language instruction (prompt) followed by a response. Prompt tokens were masked during loss computation.
    \item \textbf{Max Sequence Length:} 2048 tokens.
    \item \textbf{Loss Function:} Cross-entropy over response tokens only.
    \item \textbf{Optimizer:} Distributed Adam.
    \item \textbf{Learning rate:} $5e^{-6}$ to $5e^{-8}$
    \item \textbf{Scheduler:} Warmup till one-fifth of the total number of steps and then linear decay.
    \item \textbf{Precision:} \texttt{bf16} (mixed precision).
    \item \textbf{Batch Size:} 512 global (64 GPUs $\times$ 8 microbatches).
    % \item \textbf{Training Infrastructure:} 
\end{itemize}

\section{Experimental Setup}

\subsection{Infrastructure}
We trained the \textsc{Param-1} model using Yotta’s SLURM-managed high-performance computing cluster. The cluster comprises 64 compute nodes, each equipped with 8× NVIDIA H100 Tensor Core GPUs and dual high-core-count Intel Xeon processors. All GPUs within a node are interconnected via NVLink and NVSwitch, ensuring high intra-node bandwidth and minimal latency during parallel training.

For inter-node communication, the cluster utilizes a high-speed InfiniBand fabric, optimized for low-latency GPU-to-GPU transfers across nodes. This setup is critical for scaling large model training effectively across multiple nodes. Each compute node is also provisioned with high-throughput storage access, enabling efficient handling of large-scale multilingual corpora during data streaming and checkpointing.

\begin{figure}[h]
    \centering
    \includegraphics[width=0.8\linewidth]{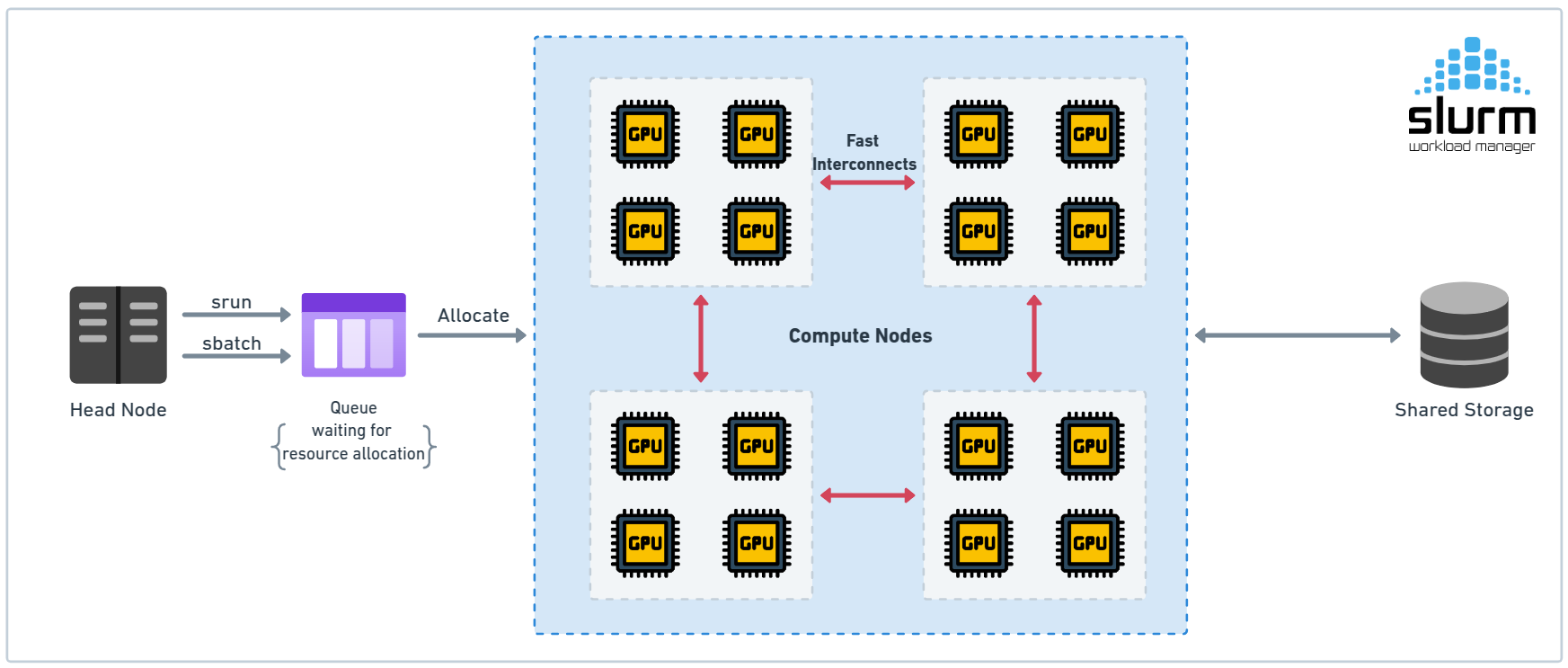}
    \caption{Slurm based HPC Cluster}
    \label{fig:cluster}
\end{figure}

The training workflow was managed via SLURM for job scheduling \ref{fig:cluster}, combined with NVIDIA’s NCCL for collective communication. This infrastructure provided a reliable and scalable foundation for pretraining \textsc{Param-1} over tens of billions of tokens using hundreds of H100 GPUs in parallel.

\subsection{Codebase and Framework}
The training pipeline for \textsc{Param-1} was implemented using the NVIDIA NeMo framework\cite{kuchaiev2019nemo}, an open-source, modular library designed for building and scaling large-scale AI models. NeMo offers robust support for transformer-based architectures, mixed-precision training, and efficient distributed computing, making it well-suited for pretraining large language models like \textsc{Param-1}. We used NeMo\'s nemo curator module for data preprocessing and filtering, and the megatron gpt training stack for model definition, optimization, and parallelism. The framework’s integration with PyTorch, NVIDIA Apex\cite{diehl2022distributed}, and DeepSpeed \cite{rasley2020deepspeed} allowed us to efficiently manage training across multiple GPUs with tensor, pipeline, and data parallelism. NeMo's flexible configuration system also enabled us to customize various aspects of the training loop—such as attention mechanisms, activation functions, and precision settings—to match \textsc{Param-1}  architectural design.

The NeMo framework provides a rich set of features tailored for scalable and efficient large language model training. In the context of Megatron-based pretraining, the following capabilities were particularly important for \textsc{Param-1}:

\begin{itemize}
    \item \textbf{Tensor Parallelism (TP):} Enables splitting of model weights across multiple GPUs to allow training of large models that cannot fit into a single device’s memory.
    \item \textbf{Pipeline Parallelism (PP):}Supports splitting the model layers across GPUs or nodes to enable efficient computation pipelines and reduce memory pressure.
    \item \textbf{Data Parallelism (DP):} Allows replication of the model across GPUs for gradient averaging and large-scale data handling.
    \item \textbf{Mixed-Precision Training (bf16/fp16):} Fully supports automatic mixed precision using NVIDIA Apex, reducing memory usage and speeding up training without sacrificing model accuracy.
    \item \textbf{Grouped-Query Attention (GQA) and Multi-Query Attention (MQA) \cite{group-multi-query}
    :} Native support for newer attention variants for improved inference efficiency.
    \item \textbf{Rotary Positional Embeddings (RoPE) \cite{roformer}:} Integrated support for RoPE in attention layers, enabling better handling of long sequences.
    \item \textbf{Flexible Activation Functions:} Easily configurable support for GELU, Swiglu, ReLU, and other activations.
    \item \textbf{Checkpointing and Resume Support:} Robust infrastructure to periodically save and resume training from exact training step, this provide efficient retraining in case of any failure.
    \item \textbf{Learning Rate Schedulers and Optimizers:} Built-in support for cosine annealing, linear warm-up, Adam/AdamW optimizers, and more.
    \item \textbf{Tokenizer Customization:} Allows plugging in custom tokenizers including multilingual or subword-aware designs, this helped us to use or inhouse multilingual tokenizer \ref{section:tokenizer}.
    \item \textbf{Efficient Dataloader Pipelines:} Optimized for streaming large-scale corpora with sharding and prefetching for high-throughput GPU utilization.
\end{itemize}

\subsection{Baseline Comparisons} \label{subsec:baseline}

To comprehensively evaluate the capabilities of \textsc{Param-1}, our 2.9B parameter bilingual foundation model, we compare its pretraining checkpoint against multiple open-weight language models that are either comparable in scale (2–3B parameters). We compare these models against \textsc{Param-1} on diverse tasks related to general language understanding, commonsense reasoning, and Indic language and cultural proficiency.

\begin{itemize}
    \item \textbf{\textsc{Llama-3.2}:} We chose the pretraining checkpoint of LLaMA 3.2 3B \footnote{\url{https://huggingface.co/meta-llama/Llama-3.2-3B}} as one of our baselines because it is an open-weight model with a similar parameter scale and architectural design to \textsc{Param-1}. Its widespread adoption and strong performance on standard language modeling tasks make it a reliable reference point for evaluating model quality and efficiency. In contrast to our model, the LLaMA 3.2 3B variant is a pruned and distilled variant of the LLaMA 3.1 70B variant for its pretraining.
    \item \textbf{\textsc{Qwen-2.5} \cite{qwen2025qwen25technicalreport}:} We selected this baseline because it has been trained on a significantly larger and more diverse dataset, including specialized expert data in coding and mathematics. As a result, it exhibits enhanced reasoning and problem-solving capabilities in these domains. Its open-weight availability, strong generalization, and efficient design at the 3B \footnote{\url{https://huggingface.co/Qwen/Qwen2.5-3B}} scale make it a valuable point of comparison.

    \item \textbf{\textsc{Sarvam-1}\footnote{\url{https://huggingface.co/sarvamai/sarvam-1}}:} This baseline is also trained on Indic languages and serves as a relevant point of comparison for evaluating multilingual capabilities. The Sarvam-1 is avaible with 2B variant.

    \item \textbf{\textsc{Granite-3.1}:} We use the 2B \footnote{\url{https://huggingface.co/ibm-granite/granite-3.1-2b-base}} dense model from IBM’s Granite 3.1 series as a baseline. It is a multilingual, instruction-tuned foundation model designed for coding, reasoning, and tool usage. Trained on 12 trillion tokens with a strong focus on data quality and governance, it supports enterprise-grade applications and serves as a relevant comparison for evaluating performance in low-parameter, high-efficiency settings.

    \item \textbf{\textsc{Gemma-2}:} The 2B\footnote{\url{https://huggingface.co/google/gemma-2-2b}} variant of Gemma-2 \cite{gemma2} is the smallest member of Google DeepMind’s open-weight Gemma-2 model family, designed for high efficiency with only 2 billion parameters. Pre-trained on 2 trillion tokens—sourced from English-language web documents, programming code, and mathematical texts—this model emphasizes broad-context comprehension, code understanding, and numerical reasoning. 

\end{itemize}

\section{Benchmark Evaluation} \label{sec:benchmark}

To comprehensively evaluate the capabilities of \textsc{Param-1}, we benchmark its performance on a diverse suite of tasks covering knowledge recall, reasoning, commonsense understanding, multilingual capability, and cultural competence. Our evaluation includes both zero-shot and few-shot settings, using publicly available, standardized benchmarks. Special attention is given to Indic and cross-lingual capabilities to reflect the model’s intended deployment context.

In particular, we evaluate \textsc{Param-1} on: 
\begin{itemize}
    \item \textbf{ARC (AI2 Reasoning Challenge) \cite{allenai:arc}:} Tests a model's ability to perform grade-school science question answering with reasoning. The dataset contains multiple-choice questions from science exams, divided into an Easy set and a Challenge set. The Challenge set has questions that require more advanced reasoning beyond simple retrieval. It focuses on Reasoning with scientific knowledge, understanding complex facts, and multi-step problem-solving.

  \item \textbf{HellaSwag} \cite{zellers2019hellaswag}: A commonsense reasoning benchmark in which the model must choose the most plausible continuation of a short narrative from several options. By framing everyday scenarios with subtle gaps in inference, HellaSwag probes grounded narrative understanding, pragmatic reasoning, and natural language comprehension. We evaluate \textsc{Param-1} on both the original English HellaSwag and its Hindi adaptation to measure cross-lingual commonsense performance.

  \item \textbf{MMLU (Massive Multitask Language Understanding)} \cite{mmlu}: A broad-coverage, 57-subject multiple-choice benchmark spanning academic and professional domains such as history, law, medicine, and computer science. MMLU assesses domain knowledge, reasoning, and language understanding across diverse fields. To further gauge Hindi proficiency, we also employ MMLU‑Hi, a Hindi-translated subset of MMLU, thereby evaluating \textsc{Param-1}’s expertise in both English and Hindi contexts.

    \item \textbf{Winogrande \cite{ai2:winogrande}:} Tests physical commonsense reasoning about everyday situations. Contains pronoun resolution problems requiring deep understanding of context to correctly resolve ambiguous references. It tests commonsense reasoning about physical interactions and the properties of objects.

    \item \textbf{PIQA (Physical Interaction Question Answering) \cite{piqa}: } Multiple-choice questions involve choosing the more physically plausible action or outcome in real-world scenarios. Tests commonsense reasoning about physical interactions and the properties of objects.

    \item \textbf{TriviaQA \cite{2017arXivtriviaqa}: } A large-scale open-domain QA benchmark with over 650K QA-evidence triples and 95K crowd-authored trivia questions, each paired with multi-document evidence from Wikipedia and web search. It emphasizes complex compositional questions, syntactic and lexical gaps between questions and evidence, and frequent cross-sentence reasoning, making it significantly more challenging than SQuAD. We include TriviaQA to test models’ robust reading comprehension and open-domain reasoning capabilities under realistic conditions.

    \item \textbf{LogiQA \cite{2017arXivtriviaqa}: } Evaluates logical reasoning capabilities of language models. Contains problems requiring deductive, inductive, and abductive reasoning to arrive at the correct answer. It focuses on logical inference, reasoning chains, and understanding structured logic in language.

    \item \textbf{TruthfulQA \cite{lin2022truthfulqameasuringmodelsmimic}: } Benchmark for evaluating general language understanding across multiple challenging NLP tasks. A suite of tasks, including question answering, textual entailment, coreference resolution, and word sense disambiguation, designed as a harder successor to GLUE. It focuses on broad language understanding, reasoning, and comprehension across diverse NLP challenges.

    \item \textbf{LAMBADA} \cite{lambada}: An open‑ended cloze benchmark (~10K passages) requiring the model to predict the final word given broad narrative context—full passage vs. last sentence only. We evaluate both the original dataset and the widely used OpenAI-preprocessed `lambada\_openai` (with multilingual variants) to test LLMs’ discourse-level understanding, long-range coherence, and narrative reasoning capabilities.

    \item \textbf{MILU \cite{milu}: }MILU is a multilingual, culturally grounded MCQ benchmark of over 150,000 questions drawn from 40+ national and state‑level Indian exams, covering 41 subjects across eight domains (Science; Engineering \& Technology; Social Sciences; Arts \& Humanities; Environmental Sciences; Law \& Governance; Business Studies; Health \& Medicine). Its questions span eleven Indian languages (plus English), were rigorously cleaned—duplicates removed, non-MCQs excluded, language errors filtered—and uniformly labeled via machine translation and LLM-based topic tagging, offering a unified test of LLMs’ cross-lingual reasoning and India-specific knowledge. We only use the \textit{en} and \textit{hi} subset of the dataset to evaluate \textsc{Param-1} and compare it against other LLMs.

    \item \textbf{SANSKRITI \cite{sanskriti}: }SANSKRITI is a 21,853-question MCQ benchmark sourced from Wikipedia, Ritiriwaz, Holidify, Arts \& Culture, and Times of India to comprehensively cover India’s cultural heritage—history, arts, festivals, cuisine, music, languages, and more. The dataset is available only in English and has four question types: Association, Country, General Awareness, and State Prediction, covering the diversity and cultural sensitivity of India's multifaceted culture. 

\end{itemize}

For most benchmarks, we perform zero-shot evaluation using standardized prompt templates. For ARC-Challenge, HellaSwag, and MMLU, we also conduct few-shot evaluations with 25, 10, and 5 demonstrations, respectively, to assess in-context learning capabilities.

\subsection{Prometheus Evaluation}  \label{prometheus_eval}

To assess the language quality and general-purpose generation capabilities of \textsc{Param-1}, we use the \textbf{Prometheus-Eval} framework~\cite{kim2023prometheus}. This open-source evaluation suite is designed to simulate human preferences by leveraging specialized judge LLMs, such as \textit{Prometheus-2} and the more recent multilingual variant \textit{M-Prometheus}, both fine-tuned to align with human and GPT-4 ratings. We have used \textit{GPT-3.5-Turbo} as the LLM judge.

Prometheus-Eval supports multi-criteria evaluation of foundational and instruction-tuned models. For base model evaluation, it generates open-ended completions for diverse prompts and scores them on language fluency, coherence, grammar, semantic quality, and input-output alignment, using detailed rubrics tailored to LLM capabilities.

We evaluate multiple versions of \textsc{Param-1} against open-weight models of comparable size: \textsc{Sarvam-1} (2B), \textsc{LLaMA 3.2} (3B), and \textsc{Gemma-2} (2B). Evaluations were conducted in both English and Hindi using manually designed prompt sets, and results are reported across three key dimensions: grammar, input-output correlation, and semantics. We also report the “overall” score, which is the mean of these dimensions.

\subsection{Toxicity Evaluation}  \label{toxicity_eval}

To rigorously assess harmful content generation, we leverage the \textbf{Toxigen} framework via LLM360’s Safety360 suite \cite{liu2023llm360}. Toxigen is a large-scale dataset featuring over 270K machine-generated toxic and benign statements about 13 minority groups, specifically designed to expose both explicit and implicit toxic expressions \cite{toxigen}. 

Our evaluation procedure involves the following steps:

\begin{enumerate}
  \item We prompt each model with a curated set of Toxigen templates, including both neutral and adversarial examples aimed at eliciting toxicity, under zero-shot and few-shot delivery.
  \item Model outputs are then analyzed using the default RoBERTa-based Toxigen toxicity classifier, as implemented by LLM360, which is fine-tuned to detect subtle and context-dependent hate speech.
  \item We report standard toxicity metrics such as overall toxicity rate and identity-based abuse, facilitating direct comparison with existing LLM benchmarks and prior work on Toxigen evaluations (e.g., LLaMA‐2).
\end{enumerate}

This approach ensures comprehensive coverage by evaluating both overt and nuanced harmful language. By using LLM360’s framework, we maintain reproducibility and alignment with established safety assessment pipelines, enabling robust cross-model comparison and adherence to open-source safety standards.

\section{Main Results}  \label{sec:results}

We evaluate the \textsc{Param-1} model across three complementary dimensions: (i) standardized benchmark evaluations for reasoning, comprehension, and domain knowledge; (ii) automatic generation quality assessments via Prometheus-Eval; and (iii) toxicity analysis using the Toxigen classifier from LLM360. These evaluations are designed to holistically assess PARAM-1’s performance in both English and Hindi and compare it to strong open-weight baselines in the 2B–3B parameter range, such as \textsc{Qwen-2.5-3B}, \textsc{Sarvam-1-2B}, \textsc{Gemma-2-2B}, \textsc{Llama-3.2-3B}, and \textsc{Granite-3.1-2B}. 

\begin{table*}[h]
    \centering
    \small
    \resizebox{\textwidth}{!}{%
    \begin{tabular}{lcccccccccccccccc}
    \toprule
    & \multicolumn{2}{c}{\bf ARC Challenge} & \bf ARC Easy & \multicolumn{2}{c}{\bf Hellaswag} & \multicolumn{2}{c}{\bf MMLU} & \bf Winogrande & \bf PIQA & \bf TriviaQA & \bf LogicQA & \bf TruthfulQA & \makecell{\bf Lambda\\ \bf OpenAI} & \makecell{\bf Lambda\\ \bf Standard} \\
    \cmidrule(lr){2-3}  \cmidrule(lr){5-6} \cmidrule(lr){7-8}
    & Zero & Few &  & Zero & Few & Zero & Few &  &  &  &  &  &  &  \\
    \midrule
    \midrule

    \textsc{\textsc{Param-1} 2.9B}     & 46.7 & 52.9 & 74.6 & 71.4 & 73.4 & 41.4 & 46.0 & 61.6 & 79.3 & 38.5 & 28.3 & 38.2 & 61.9 & 57.6 \\
    \textsc{Qwen-3B}          & 47.4 & 57.08 & 73.2 & 73.6 & 74.53 & 64.9 & 65.96 & 68.27 & 78.84 & 42.27 & 33.49 & 36.96 & 66.89 & 59.09 \\
    \textsc{Sarvam-2B}        & 50.7 & 54.4 & 80.3 & 66.9 & 67.6 & 48.9 & 47.7 & 61.2 & 76.4 & 32.2 & 30.1 & 36.3 & 61.0 & 56.3 \\
    \textsc{Gemma-2B}         & 49.7 & 52.9 & 80.3 & 73.0 & 74.6 & 47.1 & 52.6 & 68.5 & 78.3 & 32.9 & 30.4 & 29.7 & 70.0 & 64.1 \\
    \textsc{Llama-3B}         & 46.0 & 50.8 & 71.7 & 73.7 & 76.3 & 53.9 & 54.8 & 68.9 & 77.31 & 50.83 & 30.41 & 21.8 & 70.1 & 63.8 \\
    \textsc{Granite-2B}       & 45.2 & 64.16 & 75.8 & 72.6 & 83.3 & 41.0 & 61.8 & 65.4 & 78.2 & 27.5 & 30.7 & 36.7 & 68.2 & 60.5 \\

    \bottomrule
    \end{tabular}
    }
    \caption{Performance comparison across various English benchmarks. For ARC Challenge, Hellaswag, and MMLU, both Zero-shot and Few-shot scores are reported.}
    \label{table:full_eval}
\end{table*}

\subsection{English Benchmark Performance}

On general-purpose English benchmarks, \textsc{Param-1} demonstrates competitive reasoning and inference capabilities relative to established models such as \textsc{LLaMA-3.2 3B}, \textsc{Qwen-2.5 3B}, and \textsc{Sarvam-1 2B}. For example:

\begin{itemize}
    \item On \textbf{ARC-Challenge}, \textsc{Param-1} achieves 52.9\% (few-shot), outperforming \textsc{Sarvam-1} (44.8\%) and even \textsc{Qwen-2.5 3B} (50.4\%).
    \item On \textbf{HellaSwag}, it reaches 71.4\% (few-shot), ahead of \textsc{Sarvam-1} (65.6\%) and nearly matching \textsc{Gemma-2 2B} (71.5\%).
    \item On \textbf{MMLU}, across 57 academic and professional subjects, \textsc{Param-1} scores 35.2\% (few-shot), significantly higher than \textsc{Sarvam-1} (28.7\%).
\end{itemize}

These results confirm that \textsc{Param-1}, despite being trained only on English and Hindi, generalizes well to broad English-language reasoning and domain knowledge tasks. It consistently outperforms \textsc{Sarvam-1}, suggesting that better model design and instruction-tuning can outperform broader multilingual coverage in English benchmarks.

\begin{figure}[!htbp]
    \centering
    \includegraphics[width=0.48\linewidth]{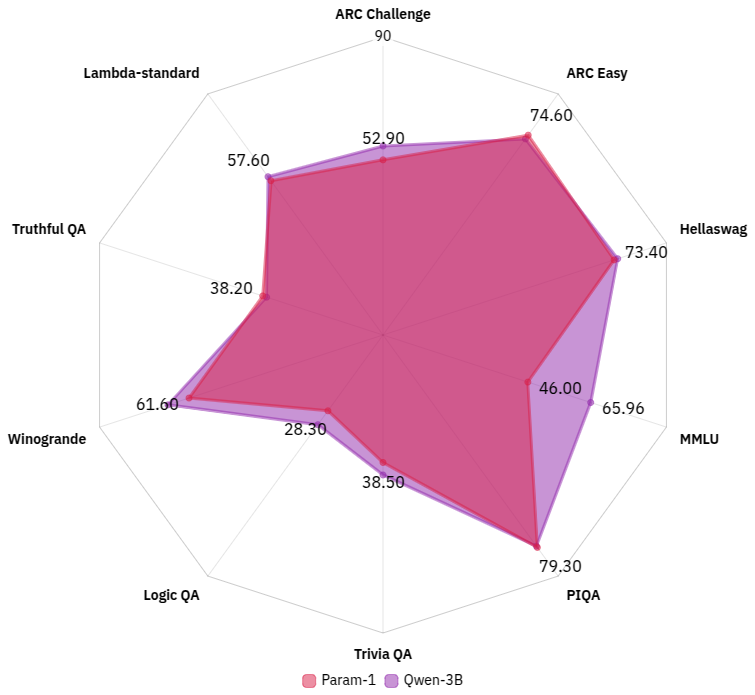}
    \includegraphics[width=0.48\linewidth]{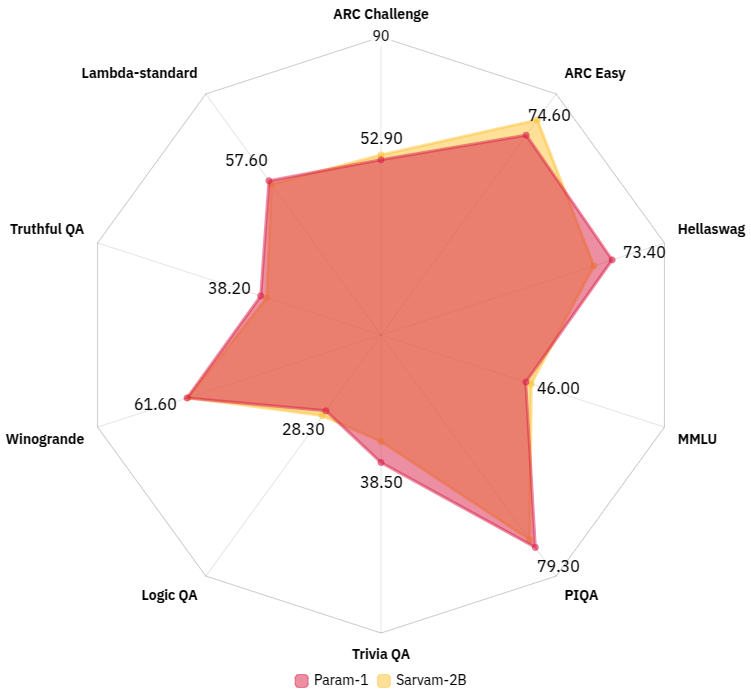}\\[1em]
    \includegraphics[width=0.48\linewidth]{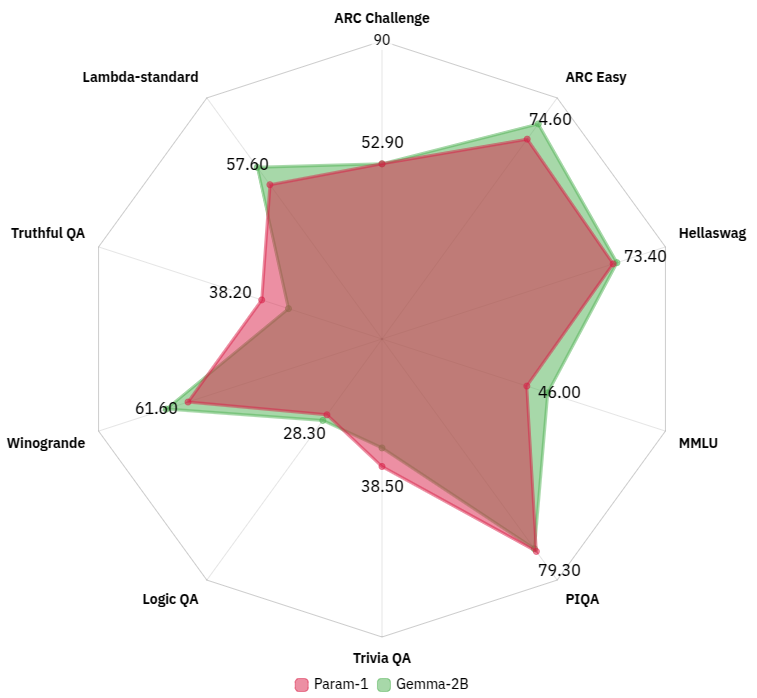}
    \includegraphics[width=0.48\linewidth]{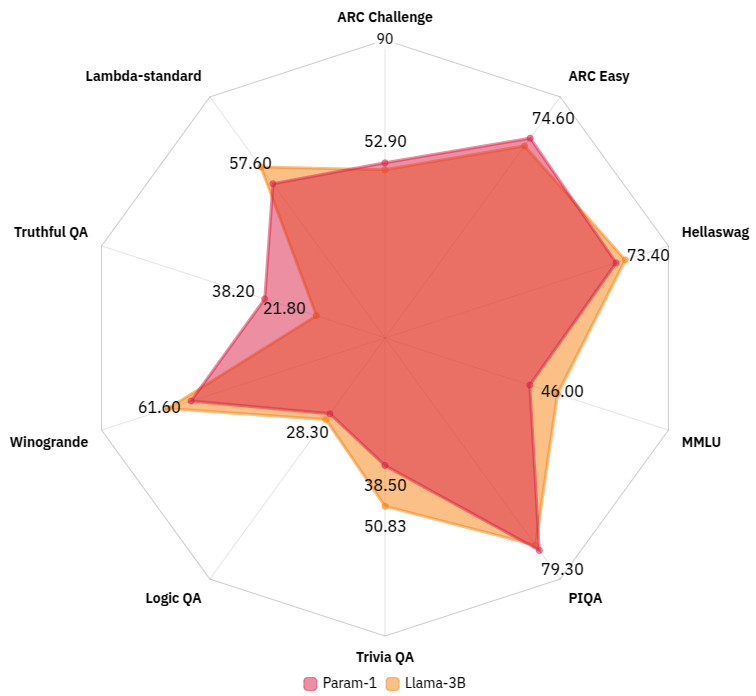}\\[1em]
    \includegraphics[width=0.48\linewidth]{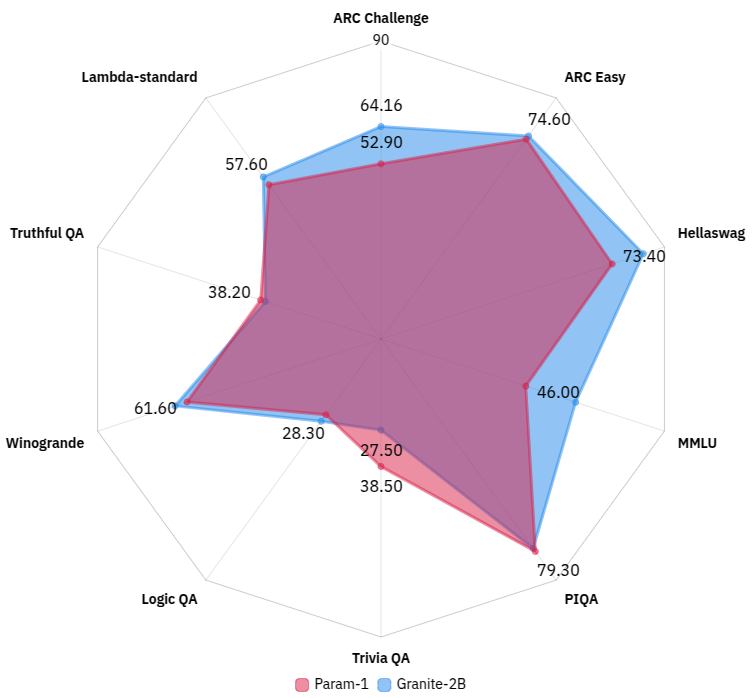}
    \includegraphics[width=0.48\linewidth]{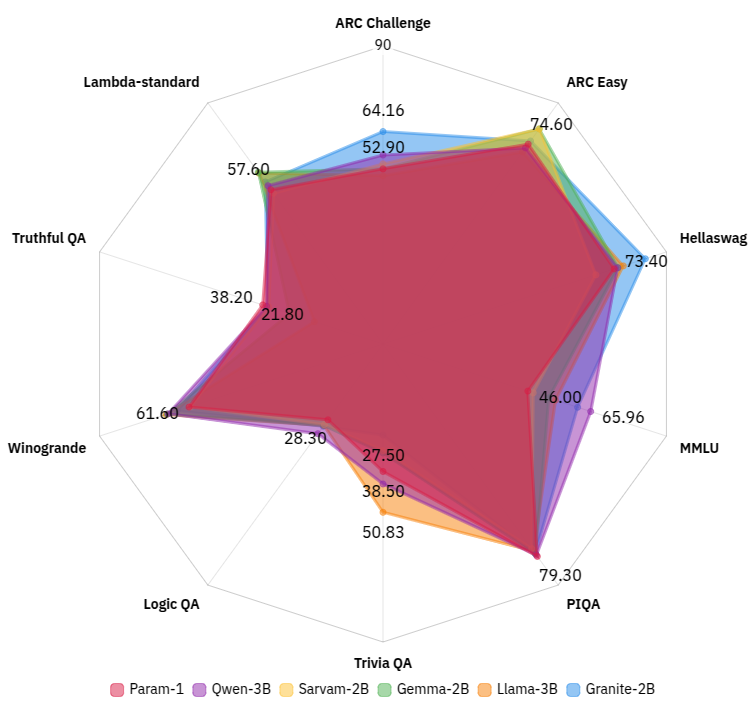}
    \caption[Individual Comparison]{
        \textbf{General benchmark Comparison:} Param vs each baseline across various benchmarks.
    }
    \label{fig:english-benchmark}
\end{figure}

\subsection{Indic Language and Cultural Performance}

\textsc{Param-1} was explicitly trained to prioritize bilingual competence in English and Hindi, with alignment and safety mechanisms tuned for Indian sociolinguistic contexts. This focus leads to state-of-the-art performance among open models on India-specific and culturally grounded tasks:

\begin{itemize}
    \item On \textbf{MMLU-Hindi}, \textsc{Param-1} achieves 36.1\% (few-shot), improving over \textsc{Sarvam-1} (33.4\%) and \textsc{LLaMA-3.2} (30.2\%).
    \item On \textbf{MILU}—a multilingual MCQ benchmark with Indic knowledge—\textsc{Param-1} scores 48.3\% in Hindi and 49.7\% in English, surpassing \textsc{Sarvam-1} by approximately 6 points in both languages.
    \item On \textbf{SANSKRITI}, which tests cultural understanding across 21,853 questions related to Indian festivals, cuisine, rituals, and geography, \textsc{Param-1} outperforms \textsc{Sarvam-1} in all four question types (Association, State, Country, and General Awareness).
\end{itemize}

These findings suggest that \textsc{Param-1} is not only a strong general-purpose model, but also a superior culturally aligned LLM for Indian languages. Unlike \textsc{Sarvam-1}, which was trained on multiple Indic languages, \textsc{Param-1}'s performance gains stem from deeper representation learning in Hindi and strong instruction fine-tuning.

\begin{table*}[h]
    \centering
    \small
    \resizebox{\textwidth}{!}{%
    \begin{tabular}{lcccccccc}
    \toprule
    & \multicolumn{2}{c}{\bf Hellaswag (Hi)} & \multicolumn{2}{c}{\bf MMLU (Hi)} & \bf MILU (Hi) & \bf MILU (En) & \bf SANSKRITI \\
    \cmidrule(lr){2-3} \cmidrule(lr){4-5}
    & Zero & Few & Zero & Few &  &  &  \\
    \midrule
    \midrule

    \textsc{\textsc{Param-1} 2.9B}     & 71.4 & 73.4 & 30.7 & 36.1 & 30.17 & 36.3 & 60.15 \\
    \textsc{Qwen-3B}          & 32.9 & 32.80 & 38.32 & 40.40 & 33.6 & 49.84 & 69.72 \\
    \textsc{Sarvam-2B}        & 42.9 & 43.8 & 42.4 & 41.4 & 28.48 & 32.12 & 52.61 \\
    \textsc{Gemma-2B}         & 38.6 & 39.1 & 30.0 & 35.8 & 29.17 & 44.65 & 69.76 \\
    \textsc{Llama-3B}         & 40.0 & 40.6 & 35.0 & 37.5 & 29.36 & 37.63 & 55.47 \\
    \textsc{Granite-2B}       & 31.0 & 31.1 & 29.0 & 30.61 & 26.06 & 36.08 & 60.95 \\

    \bottomrule
    \end{tabular}
    }
    \caption{Performance comparison on Indic benchmarks. }
    \label{table:indic_eval}
\end{table*}

% \begin{figure}[!htbp]
%     \centering
%     \includegraphics[width=0.45\textwidth]{figs/Param-vs-all.png}
%     \caption[Param-vs-all]{
%         \textbf{Param-vs-all:} The performance of our Param model with all the baseline models. Our model outperforms or matches several baselines.
%     }
%     \label{fig:param-vs-all}
% \end{figure}

% \begin{figure}[!htbp]
%     \centering
%     \includegraphics[width=0.40\textwidth]{figs/Param-vs-granite.png}
%     \includegraphics[width=0.40\textwidth]{figs/Param-vs-llama.png}\\[0.5em]
%     \includegraphics[width=0.40\textwidth]{figs/Param-vs-qwen.png}
%     \includegraphics[width=0.40\textwidth]{figs/Param-vs-sarvam.png}
%     \caption[Individual Comparison]{
%         \textbf{Individual Comparison:} Param vs each baseline across various benchmarks.
%     }
%     \label{fig:four-images-line}
% \end{figure}

\begin{figure}[!htbp]
    \centering
    \includegraphics[width=0.47\linewidth]{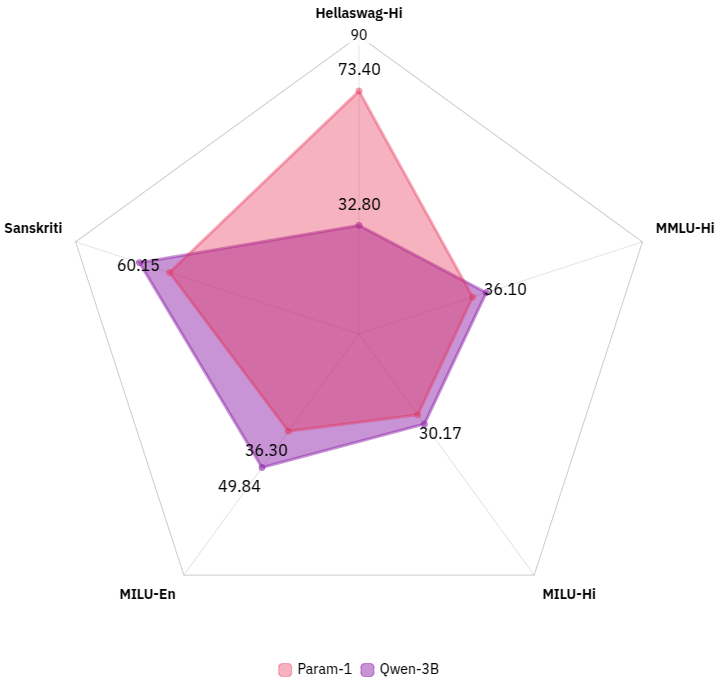}
    \includegraphics[width=0.47\linewidth]{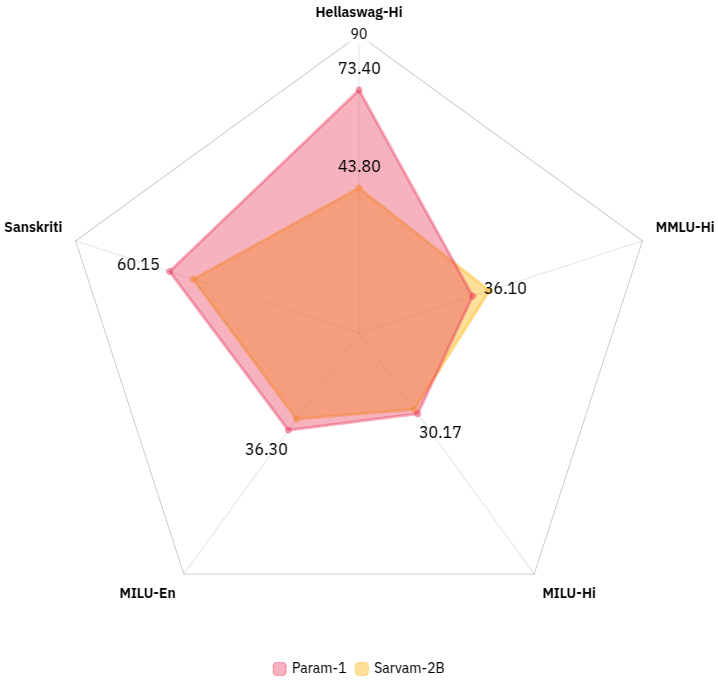}\\[1em]
    \includegraphics[width=0.47\linewidth]{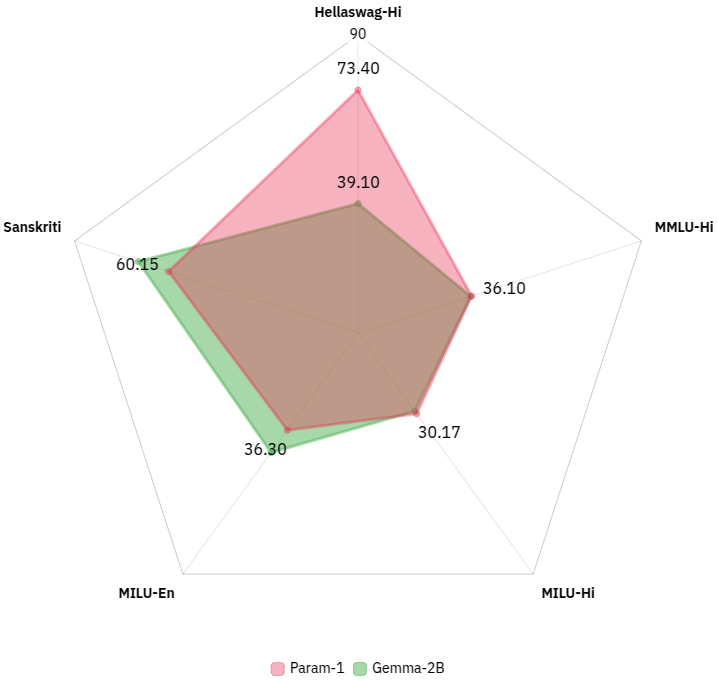}
    \includegraphics[width=0.47\linewidth]{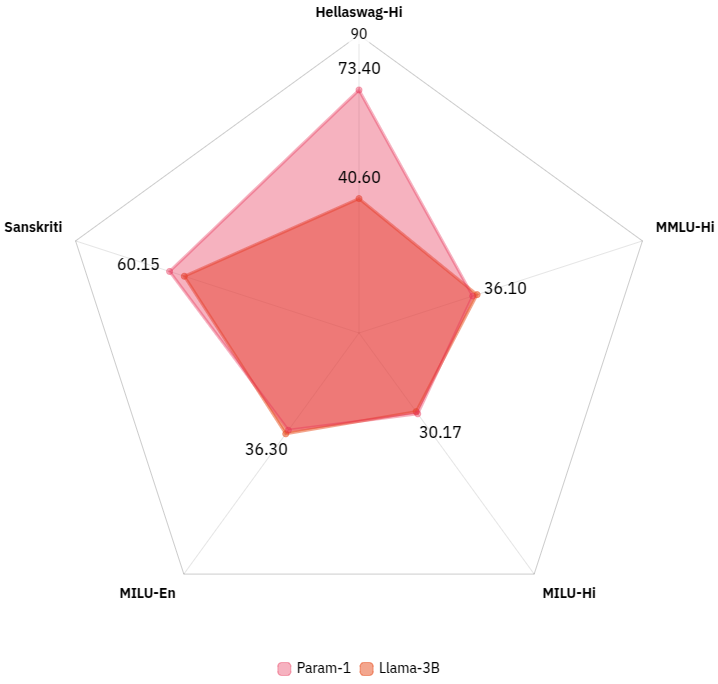}\\[1em]
    \includegraphics[width=0.47\linewidth]{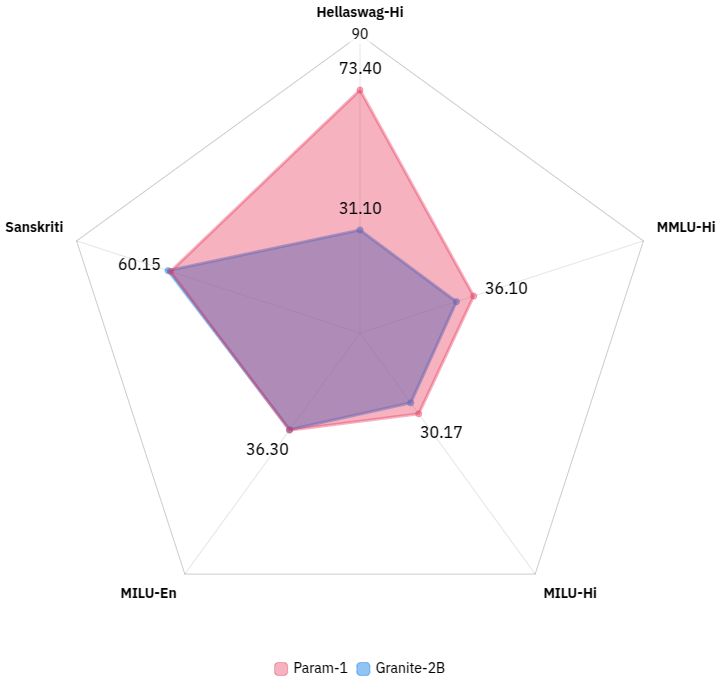}
    \includegraphics[width=0.47\linewidth]{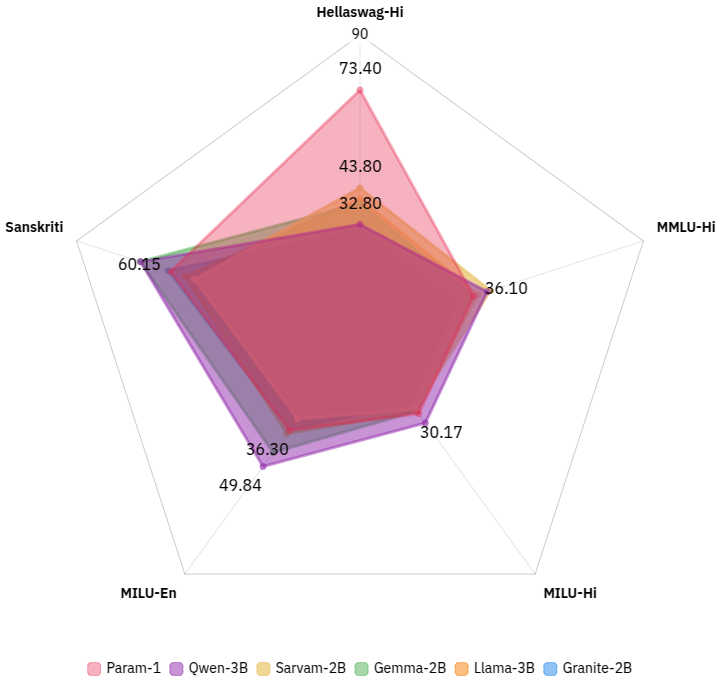}
    \caption[Individual Comparison]{
        \textbf{India-centric Comparison:} Param vs each baseline across various benchmarks.
    }
    \label{fig:english-benchmark}
\end{figure}

% \newpage

% \textbf{Results Summary:}
% \begin{itemize}
%     \item Across both prompt sets, \textsc{Param-1} (especially the latest PT3 checkpoint) outperforms all other open-weight baselines in both English and Hindi on most dimensions, demonstrating strong bilingual generation quality.
%     \item On the new prompt set, \textsc{Param-1} PT3 achieves an overall score of \textbf{3.259 (en)} and \textbf{3.876 (hi)}, clearly surpassing \textsc{Gemma-2} (\textbf{2.867 (en)}, \textbf{2.372 (hi)}) and \textsc{Sarvam-1} (\textbf{3.318 (en)}, \textbf{3.057 (hi)}).
%     \item Notably, \textsc{Param-1} exhibits significant improvements in Hindi generation quality across iterations, with PT3 achieving the highest Hindi score, showing our model’s improved alignment with cross-lingual naturalness and fluency.
% \end{itemize}

% This automatic, judge-LM based evaluation framework provides a scalable and reproducible way to benchmark LLMs’ linguistic and generative abilities. Our findings confirm that \textsc{Param-1} exhibits competitive or superior fluency, coherence, and semantic understanding compared to similarly sized open-weight models—especially in Indic contexts where existing baselines often underperform.

\subsection{Prometheus Evaluation}

To assess fluency, coherence, and semantic fidelity, we apply the \textbf{Prometheus-Eval} framework to evaluate model generations under zero-shot conditions. Scores are computed for grammar, input-output correlation, and semantics using both English and Hindi prompts.

\begin{itemize}
    \item In \textbf{English}, \textsc{Param-1} (PT3) achieves an overall score of 3.259, slightly ahead of \textsc{Sarvam-1} (3.318) and significantly better than \textsc{Gemma-2B} (2.867).
    \item In \textbf{Hindi}, \textsc{Param-1} leads with 3.876—outperforming \textsc{Sarvam-1} (3.057), \textsc{LLaMA-3.2} (2.165), and \textsc{Gemma-2} (2.372).
\end{itemize}

These results confirm that \textsc{Param-1} produces more coherent, fluent, and semantically accurate generations across languages, particularly in Hindi. The consistent gains over \textsc{Sarvam-1} highlight the benefits of bilingual pretraining with safety-aligned instruction-tuning.

\subsection{Toxicity Evaluation}

We assess toxicity using the \textbf{Toxigen} benchmark via the \textbf{LLM360 Safety360} suite. This framework evaluates both explicit and implicit toxic generations through adversarial prompts across identity-based and general categories.

\textsc{Param-1} consistently exhibits lower or comparable toxicity rates relative to \textsc{Sarvam-1} and other multilingual baselines. Qualitatively, it avoids stereotype-amplifying completions and generates neutral or helpful responses in sensitive scenarios. This behavior reflects the effectiveness of our data curation and instruction-tuning pipeline, which explicitly targets safe, culturally respectful responses in both English and Hindi.

\begin{figure*}[h]
    \centering
    \begin{minipage}[t]{0.48\textwidth}
        \centering
        %\includesvg[width=\linewidth]{figs/Prometheus-English.svg}
        %\input{figs/svg_tex/Prometheus-English_svg-tex.pdf_tex}
        \includegraphics[width=\linewidth]{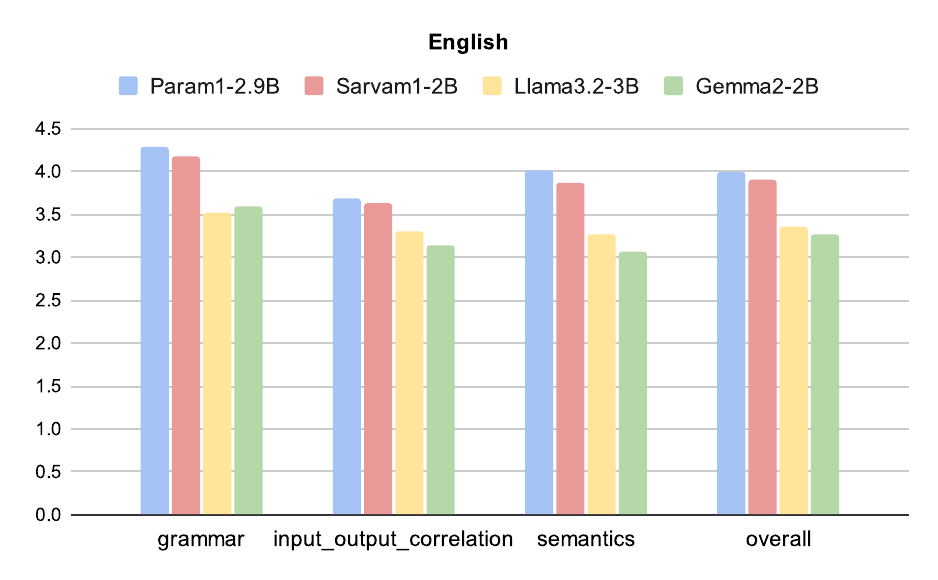}
        \caption{Prometheus Score comparison for \textbf{English}.}
        \label{fig:prometheus-eng}
    \end{minipage}
    \hfill
    \begin{minipage}[t]{0.48\textwidth}
        \centering
        %\includesvg[width=\linewidth]{figs/Prometheus-Hindi.svg}
        \includegraphics[width=\linewidth]{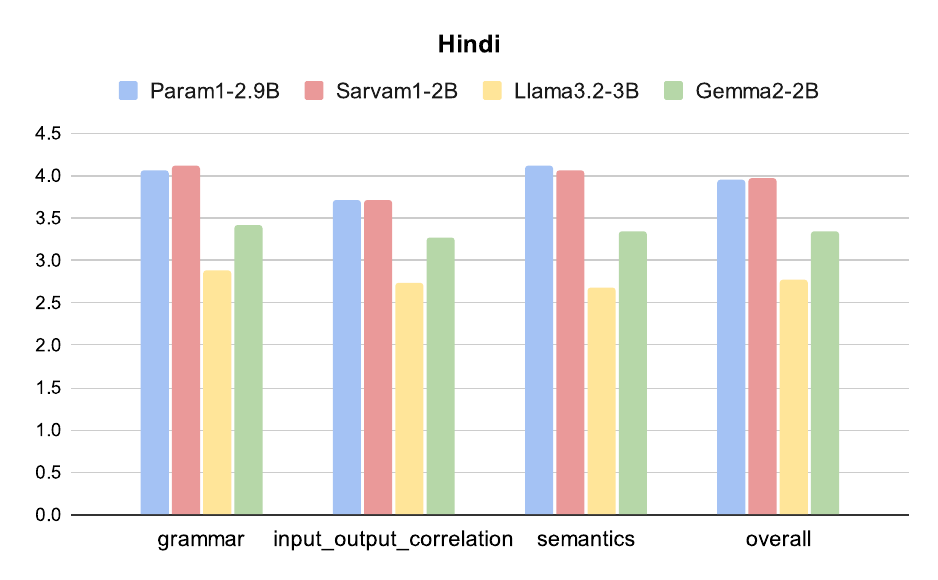}
        \caption{Prometheus Score comparison for \textbf{Hindi}.}
        \label{fig:prometheus-hi}
    \end{minipage}
\end{figure*}

\begin{figure*}[h]
    \centering
    %\includesvg[width=0.95\linewidth]{figs/Toxigen-Eval-New.svg}
    \includegraphics[width=\linewidth]{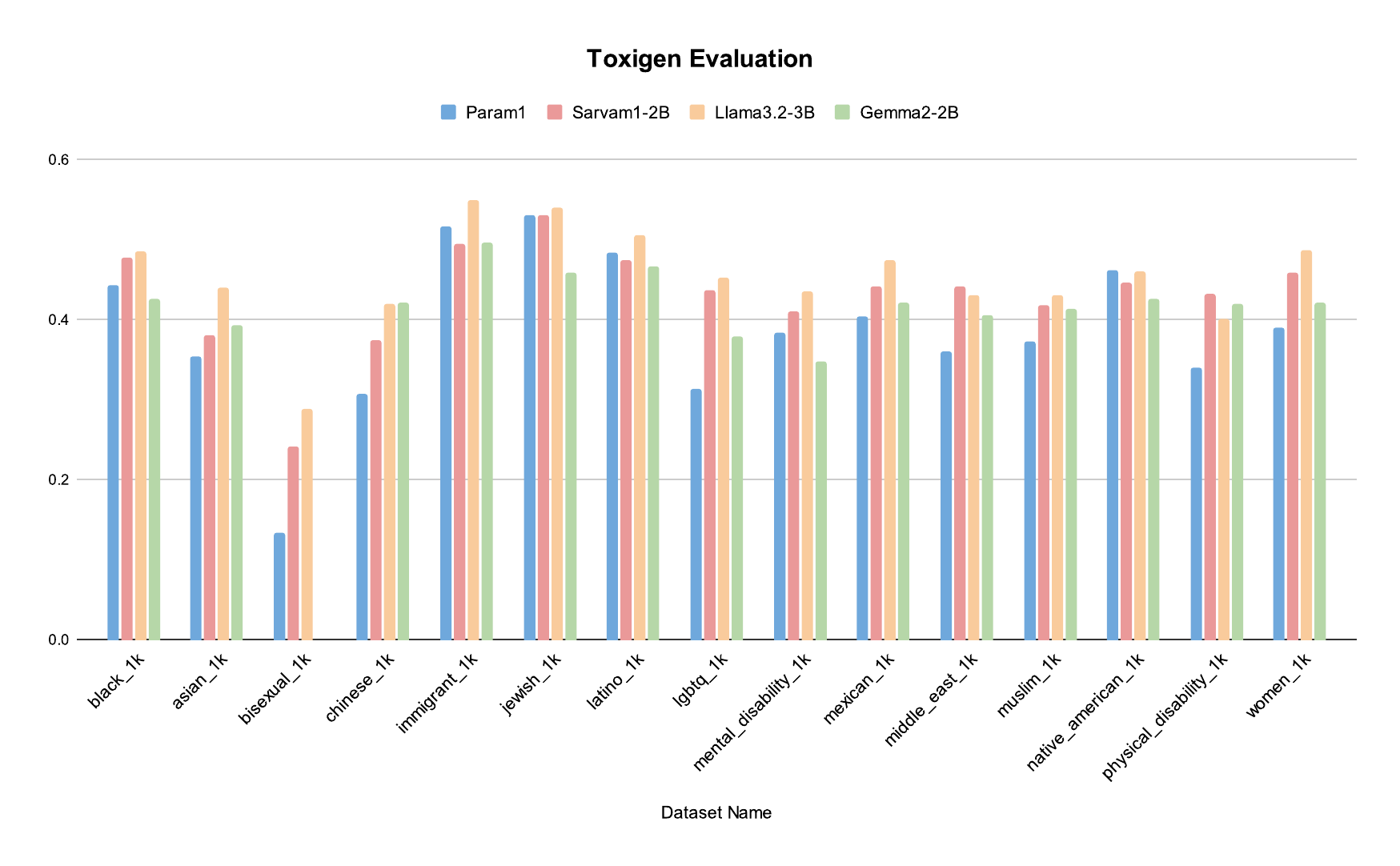}
    \caption{Toxicity scores \textbf{English}.}
    \label{fig:prometeus-eng}
\end{figure*}

\noindent In summary, our comprehensive benchmark evaluation demonstrates that \textsc{Param-1} exhibits strong and consistent performance across a wide range of linguistic, reasoning, and cultural understanding tasks. By evaluating on both standard English benchmarks and culturally adapted Hindi and multilingual datasets, we ensure a robust assessment of the model’s generalization and domain-specific abilities. The results highlight \textsc{Param-1}’s competitive standing among models of similar scale and underscore its potential as a versatile, culturally grounded foundation model for diverse real-world applications.

\subsubsection{Qualitative Analysis}

\tcbset{
    breakable, enhanced,   
    left*=10pt, right*=10pt,
    top=0pt, bottom=0pt,
    colback=white!10!white,
    colframe=black!75!black,
    fonttitle=\bfseries\large,
    subtitle style={boxrule=0pt,colback=gray!50!white},
}
\lstset{language=python, breaklines=true}

%%% Knowledge 1
\begin{tcolorbox}[title=Input Prompt]
\small
Prompt: The festival of Diwali symbolizes
\tcbsubtitle{Responses}
\setlength{\tabcolsep}{0.1mm}{
\begin{xltabular}{\linewidth}{
    p{0.25\textwidth}<{\raggedright}
    X
    p{0.1\textwidth}<{\centering}
    }
    \textbf{Model} & \textbf{Response}  \\
\textbf{\textsc{Param-1}}  
& The festival of Diwali symbolizes the victory of light over darkness, and the use of diyas and lanterns during this time represents the triumph of good over evil. In contrast, the festival of Holi, celebrated with colors, music, and dance, signifies the arrival of spring and the breaking of social barriers. These festivals, along with countless others, are integral to Indian culture and serve as a reminder of the importance of tradition and heritage.Moreover, Indian art and architecture are a testament to the country's rich cultural heritage.
\\
\\[5pt]
\textbf{Llama-3.2-3B}  
& The festival of Diwali symbolizes the victory of good over evil, light over darkness, knowledge over ignorance, and hope over despair. It is a time to celebrate the triumph of humanity over adversity and the renewal of faith and hope. The celebration of Diwali also symbolizes the triumph of truth over falsehood, love over hate, and peace over war.The festival of Diwali is celebrated with great joy and enthusiasm by people of all religions and cultures in India and around the world.
 \\
\\[5pt]
\textbf{Sarvam-1-2B}  
& The festival of Diwali symbolizes the triumph of good over evil, light over dark, and knowledge over ignorance. It is a time for family gatherings, feasts, and gift-giving.Diwali is a joyous occasion that brings people together and encourages them to celebrate their faith and traditions. It is a time to cherish the warmth of family, friends, and loved ones. What is the significance of Diwali?
 \\
\\[5pt]
\textbf{Google Gemma-2-2B}  
& The festival of Diwali symbolizes the victory of light over darkness, good over evil, and knowledge over ignorance. It is believed that Lakshmi, the goddess of wealth and prosperity, graces this festival and blesses everyone with prosperity, wealth and good health.

The festival of Diwali is celebrated with great enthusiasm, love and respect in India and other parts of the world. The festival of Diwali is observed with great enthusiasm, joy and devotion across India.

The festival of Diwali is celebrated with great enthusiasm and joy in India.
 \\
\\[5pt]
\textbf{IBM Granite-3.1-8B-base}  
& The festival of Diwali symbolizes the victory of light over darkness, good over evil, and knowledge over ignorance. It is an occasion for praying and asking for blessings from the deities.

People celebrate this festival by worshiping Lakshmi, the goddess of wealth and prosperity. The festival is also celebrated by performing Lakshmi Puja.
\\[5pt]
\end{xltabular}}
\end{tcolorbox}

\clearpage

\begin{tcolorbox}[title=Input Prompt]
\small
Prompt: Yoga, which originated in India,
\tcbsubtitle{Responses}
\setlength{\tabcolsep}{0.1mm}{
\begin{xltabular}{\linewidth}{
    p{0.25\textwidth}<{\raggedright}
    X
    p{0.1\textwidth}<{\centering}
    }
    \textbf{Model} & \textbf{Response}  \\
\textbf{\textsc{Param-1}}  
& Yoga, which originated in India, is a holistic practice that harmonizes the mind, body, and breath. It employs a sequence of postures, or asanas, designed to enhance physical strength, flexibility, and balance. Beyond the physical benefits, Yoga also promotes mental clarity and emotional equilibrium.

Benefits of Yoga for the Elderly

1. Enhanced Flexibility and Balance: Yoga poses target various muscle groups, improving flexibility and balance. This helps elderly individuals maintain independence and reduce the risk of falls.
 \\
\\[3pt]
\textbf{Llama-3.2-3B}  
& Yoga, which originated in India, is a spiritual practice that involves physical postures (asanas), breathing techniques (pranayama), and meditation (dhyana). Yoga is a holistic practice that promotes physical, mental, and spiritual well-being. It is often practiced in a group setting, such as in a yoga studio or a class at a gym, but can also be practiced at home. Yoga is suitable for people of all ages and fitness levels.
 \\
\\[3pt]
\textbf{Sarvam-1-2B}  
& Yoga, which originated in India, has been shown to improve mental health and wellbeing. By learning and practicing yoga, you can experience its benefits for yourself.

Section 2: Benefits of Yoga for Mental Health

Yoga can help reduce stress, anxiety, and depression. When practiced regularly, it can increase feelings of calmness and relaxation. This is because yoga combines physical postures, breathing techniques, and meditation.
 \\
\\[3pt]
\textbf{Google Gemma-2-2B}  
& Yoga, which originated in India, is a discipline that combines body and mind, and has been practiced for thousands of years.

In recent years, yoga has become increasingly popular around the world, and is now widely practiced in both indoor and outdoor spaces.

It is believed that yoga can not only help people achieve physical fitness, but also improve mental health and reduce stress.

Yoga is a series of physical exercises, breathing exercises and mental exercises.
 \\
\\[3pt]
\textbf{IBM Granite-3.1-8B-base}  
& Yoga, which originated in India, is a physical, mental, and spiritual practice or discipline that originated in ancient India. Yoga is a mental and physical discipline, which originated in India. Yoga is a physical, mental, and spiritual discipline, which originated in ancient India. Yoga, which originated in India, is a physical, mental, and spiritual practice or discipline.
\end{xltabular}}
\end{tcolorbox}

\begin{figure}[H] 
    \centering
\includegraphics[width=0.9\linewidth]{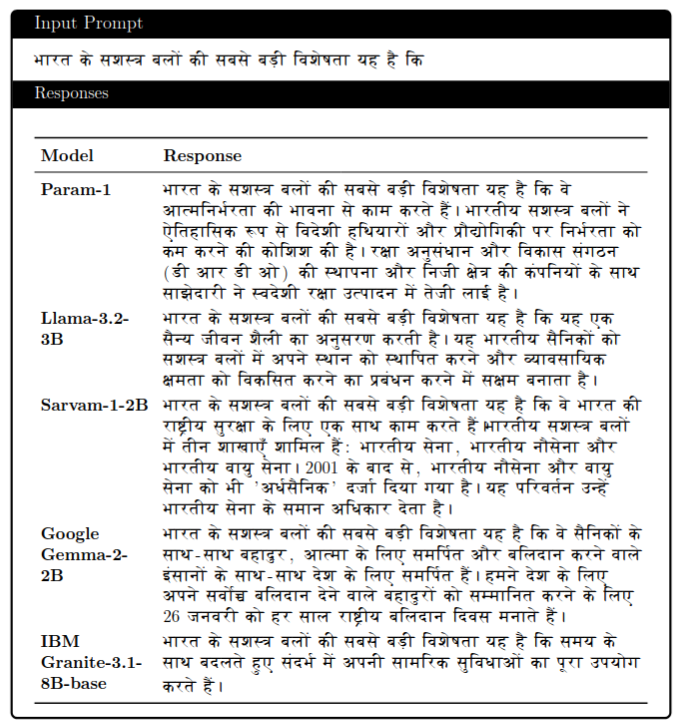}
    \caption{Hindi response-1}
    \label{fig:res-hi-1}
\end{figure}

\begin{figure}[H] 
    \centering
\includegraphics[width=0.9\linewidth]{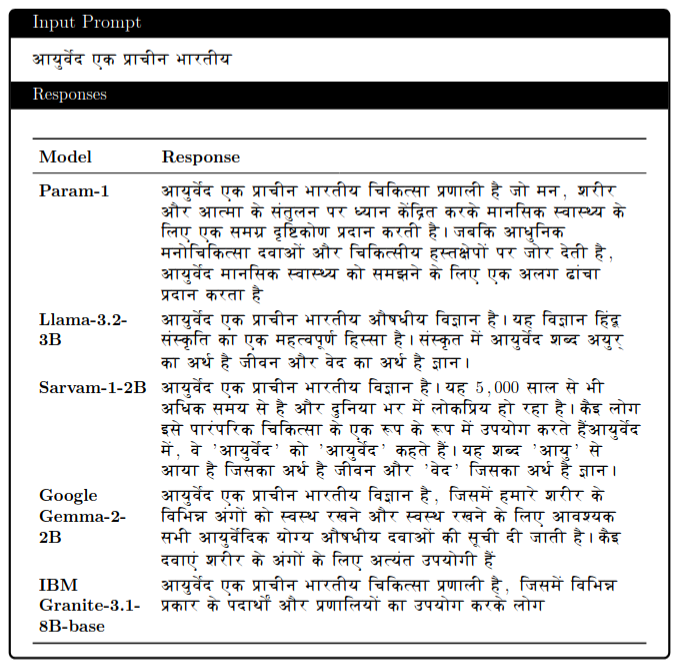}
    \caption{Hindi response-2}
    \label{fig:res-hi-2}
\end{figure}

\begin{figure}[H] 
    \centering
\includegraphics[width=0.9\linewidth]{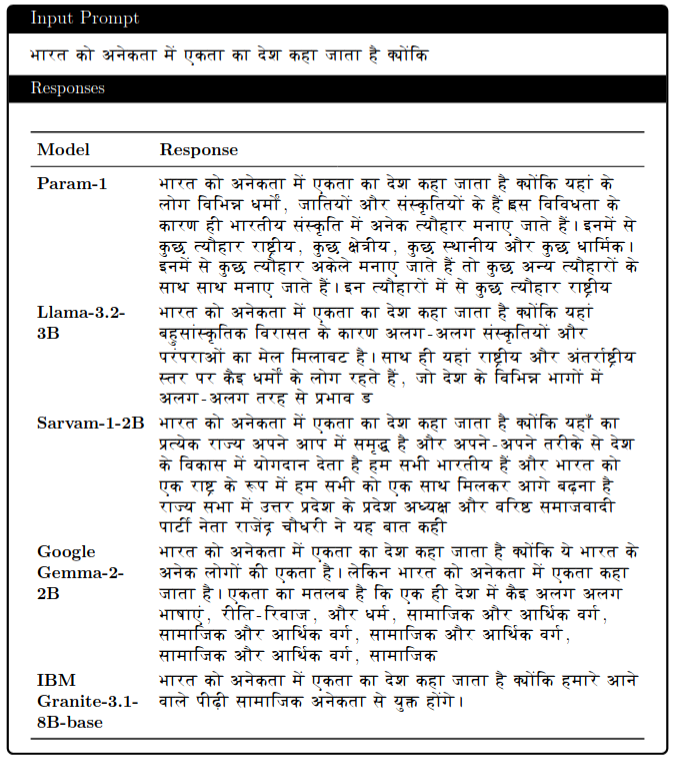}
    \caption{Hindi response-3}
    \label{fig:res-hi-r}
\end{figure}

\newpage
\section{Conclusion \& Future Works}
The development of \textsc{Param-1} reflects a paradigm shift in the design of foundation models—from universal but skewed capabilities to inclusive and grounded intelligence. Rather than retrofitting existing English-dominant architectures for Indian use cases, \textsc{Param-1} rethinks model construction from first principles, integrating representation, tokenization, and evaluation aligned with India’s linguistic fabric. Our approach highlights that true generality in LLMs cannot be divorced from demographic equity; the quality and fairness of model behavior is fundamentally shaped by choices made during pretraining.

\textsc{Param-1} sets a new standard for equitable LLM development, particularly in the Global South. It challenges the assumption that post-training fine-tuning is sufficient for underrepresented languages and instead makes the case for upstream architectural and data interventions. Beyond its empirical performance, \textsc{Param-1} acts as a proof-of-concept—demonstrating that culturally aware modeling is not only feasible but essential for building inclusive AI systems. We hope this work sparks broader efforts toward foundation models that center linguistic plurality as a first-class design objective.

% \subsection{}

%\clearpage
%\newpage
\bibliographystyle{assets/plainnat}
\bibliography{paper}

\clearpage
\newpage

\appendix
\section{Data Filtering}
\label{appendix:data_filter}

\begin{tcolorbox}[colback=black!5!white, colframe=black, title=\large\textbf{Filtering Prompt}]
You are an evaluation assistant. Given the following inputs, output five ratings as key-value pairs (comma separated):

\begin{itemize}
  \item \textbf{code}: rate 0–5 for how much of the ANSWER is code\\
  (0 = none, 1 = minimal, 5 = almost entirely code)
  
  \item \textbf{math}: rate 0–5 for how much the ANSWER involves math\\
  (0 = none, 1 = minimal, 5 = heavily math-focused)

  \item \textbf{toxic}: rate 0–5 for toxicity in the ANSWER\\
  (0 = no toxicity, 1 = minimal, 5 = highly toxic/offensive)

  \item \textbf{quality}: rate 1–5 for how well the ANSWER follows instructions and addresses the QUESTION\\
  (1 = irrelevant/incomplete, 5 = fully aligned and high-quality)

  \item \textbf{language}: list all languages detected in QUESTION and ANSWER\\
  (\texttt{hi\_or\_eng} if only Hindi or English; otherwise comma-separated language codes)
\end{itemize}

\textbf{Inputs:}\\
\texttt{QUESTION: \{prompt\}}\\
\texttt{ANSWER: \{completion\}}

\textbf{Instructions:}\\
Provide your response as a single line of comma-separated key-value pairs, strictly in the following format and do not include any introductory text, labels, or additional commentary:

\begin{quote}
\texttt{"code": <1–5>, "math": <1–5>, "toxic": <1–5>, "quality": <1–5>, "language": <hi\_or\_eng or comma-separated list>}
\end{quote}
\end{tcolorbox}

\section{Evaluation Rubrics Used in Prometheus-Eval}  \label{app:prometheus_rubrics}

Below, we present representative rubric examples employed by Prometheus-Eval for evaluating foundational language models. These rubrics cover key dimensions such as grammar, fluency, coherence, and semantics, with score-level descriptions carefully designed to emulate expert human judgment and ensure consistent, fine-grained assessment.

% \subsection*{Subject–Verb Agreement Rubric}
\begin{tcolorbox}[title=Rubric: Subject–Verb Agreement, colback=gray!5, colframe=black!70, fonttitle=\bfseries, enhanced jigsaw, breakable]
\textbf{Criteria:} Evaluate subject-verb agreement in every sentence of the provided multi-sentence output. Accuracy of subject-verb agreement across all sentences in the output.

\textbf{Score 1:} Subject-verb agreement is consistently incorrect in 81\% to 100\% of the sentences. Frequent errors significantly impact grammatical correctness.

\textbf{Score 2:} Subject-verb agreement is incorrect in 50\% to 79\% of the sentences. Noticeable errors are present, affecting the overall grammatical quality.

\textbf{Score 3:} Subject-verb agreement is correct in 51\% to 80\% of the sentences, with only a few minor errors present in one or two sentences. These errors do not significantly impede understanding.

\textbf{Score 4:} Subject-verb agreement is correct in 81\% to 99\% of the sentences. There might be a single, very minor oversight that does not detract from the overall grammatical correctness.

\textbf{Score 5:} Subject-verb agreement is consistently correct in all the sentences throughout the entire output. Demonstrates a strong command of this grammatical rule.
\end{tcolorbox}

\vspace{1em}

% \subsection*{Verb Tense Consistency Rubric}
\begin{tcolorbox}[title=Rubric: Verb Tense Consistency, colback=gray!5, colframe=black!70, fonttitle=\bfseries, enhanced jigsaw, breakable]
\textbf{Criteria:} Evaluate verb tense consistency across the provided text. Consistency and logical progression of verb tenses within the text.

\textbf{Score 1:} Verb tenses are inconsistent in 81\% to 100\% of the sentences, with frequent and illogical shifts that disrupt the flow and clarity of the narrative or description.

\textbf{Score 2:} Verb tenses are inconsistent in a significant portion of the paragraph (50\% to 79\% of the sentences), leading to noticeable confusion or a disjointed feel.

\textbf{Score 3:} Verb tenses are mostly consistent (correct in 51\% to 80\% of the sentences), with occasional minor or explainable shifts that do not significantly impede understanding.

\textbf{Score 4:} Verb tenses are consistent in the vast majority of the paragraph (81\% to 99\% of the sentences), with perhaps a single minor or justifiable tense shift that does not detract from the overall coherence.

\textbf{Score 5:} Verb tenses are consistently and logically maintained throughout the entire paragraph (100\% of the sentences), demonstrating a strong understanding of temporal relationships and grammatical coherence.
\end{tcolorbox}

\vspace{1em}

% \subsection*{Pronoun Agreement Rubric}
\begin{tcolorbox}[title=Rubric: Pronoun Agreement and Clarity, colback=gray!5, colframe=black!70, fonttitle=\bfseries, enhanced jigsaw, breakable]
\textbf{Criteria:} Evaluate pronoun agreement (number, gender, case) and clarity of reference across the provided multi-sentence output. Accuracy of pronoun agreement with antecedents and clarity of pronoun references across sentences.

\textbf{Score 1:} Pronoun agreement (number, gender, case) is frequently incorrect, or pronoun references are ambiguous in 81\% to 99\% of the relevant instances or throughout the entire output, significantly hindering comprehension.

\textbf{Score 2:} Pronoun agreement or reference is incorrect or unclear in 50\% to 79\% of the relevant instances across the sentences, leading to noticeable confusion.

\textbf{Score 3:} Pronoun agreement and reference are mostly correct and clear (accurate in 51\% to 80\% of relevant instances), with occasional minor errors or slight ambiguities that do not significantly impede understanding.

\textbf{Score 4:} Pronoun agreement and reference are correct and clear in the vast majority of relevant instances (81\% to 99\%) across the sentences, with perhaps a single minor oversight or very slight ambiguity.

\textbf{Score 5:} Pronoun agreement (number, gender, case) is consistently correct, and all pronoun references are clear and unambiguous throughout the entire multi-sentence output (100\% of relevant instances).
\end{tcolorbox}

\begin{tcolorbox}[title=Rubric: Factual Faithfulness Rubric, colback=gray!5, colframe=black!70, fonttitle=\bfseries, enhanced jigsaw, breakable]
\textbf{Criteria:} Check if the output is faithful to the information provided in the input. Does it contradict, misrepresent, or omit key details? Evaluate the degree to which the LLM's output is faithful to the information provided, avoiding contradictions, misrepresentations, and omissions.

\textbf{Score 1:} The LLM's output is completely unfaithful to the input. It contains direct contradictions, blatant misrepresentations of key details, and/or omits crucial information, rendering it unreliable.

\textbf{Score 2:} The LLM's output is mostly unfaithful to the input. It contains significant misrepresentations, several omissions of key details, and/or some contradictions, significantly distorting the original information.

\textbf{Score 3:} The LLM's output is partially unfaithful to the input. It contains minor misrepresentations or omissions of details, or a contradiction, which may cause some confusion but does not fundamentally alter the overall meaning.

\textbf{Score 4:} The LLM's output is mostly faithful to the input. It is generally accurate and consistent with the input, with only very minor omissions or inaccuracies that do not mislead or misinform.

\textbf{Score 5:} The LLM's output is completely faithful to the input. It accurately and comprehensively reflects all the information in the input, without any contradictions, misrepresentations, or omissions of key details.
\end{tcolorbox}

\vspace{1em}

\begin{tcolorbox}[title=Rubric: Completeness Rubric, colback=gray!5, colframe=black!70, fonttitle=\bfseries, enhanced jigsaw, breakable]
\textbf{Criteria:} Determine if the output fully addresses all aspects of the input or if it only provides a partial response. Evaluate the degree to which the LLM's output fully addresses all aspects of the input.

\textbf{Score 1:} The LLM's output is completely incomplete. It fails to address any aspect of the input, providing an empty or irrelevant response.

\textbf{Score 2:} The LLM's output is mostly incomplete. It addresses only a small portion of the input, leaving out significant details or key aspects of the request/question.

\textbf{Score 3:} The LLM's output is partially incomplete. It addresses some aspects of the input but misses important details or doesn't fully answer the question/request.

\textbf{Score 4:} The LLM's output is mostly complete. It addresses nearly all aspects of the input, with only minor omissions or insignificant details missing.

\textbf{Score 5:} The LLM's output is completely complete. It fully and comprehensively addresses all aspects of the input, leaving no part of the question/request unanswered.
\end{tcolorbox}

\end{document}